\documentclass{article}

\usepackage{microtype}
\usepackage{graphicx}
\usepackage{dblfloatfix}
\usepackage{subcaption}
\usepackage{booktabs} 
\usepackage[most]{tcolorbox}   
\usepackage{xcolor}            
\usepackage{colortbl}         
\usepackage{cuted}  
\usepackage{booktabs}
\usepackage{tabularx}
\usepackage{float}
\usepackage{graphicx}
\usepackage{booktabs}    
\usepackage{adjustbox}   
\usepackage{multirow}
\usepackage{caption}
\usepackage{cuted}
\usepackage{xcolor}
\usepackage{soul}
\usepackage{placeins}

\definecolor{baselinecolor}{RGB}{255,230,230}
\definecolor{sftcolor}{RGB}{230,240,255}
\definecolor{rlcolor}{RGB}{230,255,230}

\newcommand{\baselinehl}[1]{\sethlcolor{baselinecolor}\hl{#1}}
\newcommand{\sfthl}[1]{\sethlcolor{sftcolor}\hl{#1}}
\newcommand{\rlhl}[1]{\sethlcolor{rlcolor}\hl{#1}}

\usepackage{tikz}
\usetikzlibrary{arrows.meta, positioning, shapes.geometric, fit}

\usepackage[preprint]{icml2026}

\usepackage{amsmath}
\usepackage{enumitem}
\usepackage{amssymb}
\usepackage{mathtools}
\usepackage{amsthm}
\usepackage{booktabs}
\usepackage{cuted}
\usepackage{caption}
\usepackage{booktabs}
\usepackage{xcolor}
\usepackage{adjustbox}
\usepackage{tcolorbox}
\usepackage{soul}

\usepackage{tabularx}
\usepackage{booktabs}
\usepackage{array}

\definecolor{ImpactGreen}{RGB}{225,245,228}
\definecolor{ImpactBlue}{RGB}{225,235,250}
\definecolor{ImpactRed}{RGB}{250,230,230}

\definecolor{darkgreen}{RGB}{0,120,40}
\definecolor{darkblue}{RGB}{0,90,160}

\usepackage[capitalize,noabbrev]{cleveref}

\theoremstyle{plain}

\theoremstyle{definition}

\theoremstyle{remark}

\usepackage[textsize=tiny]{todonotes}

\icmltitlerunning{Learning to Ideate for Scientific Impact}

\begin{document}

\twocolumn[
 
\icmltitle{Learning to Ideate for Scientific Impact}
  \icmlsetsymbol{equal}{*}

\begin{icmlauthorlist}
\icmlauthor{Shubham Kale}{tcs}
\icmlauthor{Aniketh Garikaparthi}{tcs}
\icmlauthor{Manasi Patwardhan}{tcs}
\end{icmlauthorlist}

\icmlaffiliation{tcs}{TCS Research, India}

\icmlcorrespondingauthor{Shubham Kale}{shubham.kale4@tcs.com}

\icmlkeywords{Impact Alignment, Reinforcement Learning, Scientific Forecasting, Scientific Ideation, Reward Modeling, Large Language Models}

\vskip 0.3in
]

\printAffiliationsAndNotice{}

\begin{abstract}
    Scientific ideation is increasingly mediated by large language models, but current ideation systems are usually trained and evaluated on immediately judgeable proxies such as novelty, clarity, and feasibility. This leaves open whether delayed signals of scientific uptake can be used as feedback for steering models toward research directions with higher expected \emph{impact}. We study this question using citation-normalized impact as a noisy but scalable proxy for scholarly uptake. We construct a large-scale dataset from over 100K computer science papers by extracting goal-conditioned idea descriptions and assigning each paper an ordinal, year-normalized citation label. We then train a goal-conditioned reward model to predict citation-impact labels from research goal and idea pairs, and use this reward to align an idea generator through supervised fine-tuning followed by reinforcement learning. To reduce circularity, we evaluate generated ideas with a held-out, reference-grounded protocol that compares model outputs against historical ideas under the same research goal and weights judgments by the reference idea’s citation-impact label.  Experiments show that our RL-tuned model consistently produces ideas with higher estimated impact than both the base model and supervised fine-tuning baselines. Our findings position scientific impact as a practical, outcome-grounded feedback signal for aligning LLMs in open-ended scientific discovery.
\end{abstract}

\section{Introduction}

The onset of large language models (LLM) use in science has led to an asymmetric rise in the number of publications, without substantial evidence for increase in quality at a comparable rate \cite{Birhane2023ScienceIT, Messeri2024ArtificialIA, kusumegi_sci_2026}.
Within this shift, LLMs have expanded their role in scientific discovery from writing assistance to broader support across the research life cycle, including literature understanding, hypothesis generation, experimentation, and end‑to‑end research automation \cite{zhang2024scientificllms, zhang2025scientificmethod, lu2026automation}.
Notably, research ideation has emerged as a key driver in adoption where systems such as SciMON, ResearchAgent, SciMuse, DeepInnovator, and IRIS show that LLMs can generate, refine, and evaluate literature-grounded ideas through iterative search, expert evaluation, and human‑in‑the‑loop interaction \cite{wang2024scimon, baek2025researchagent, gu2025scimuse, fan2026deepinnovator, garikaparthi_iris}.  

However, these systems largely optimize for \emph{proxy} criteria such as novelty, feasibility or clarity over downstream impact \cite{li2024ldc, guo2024ideabench, qiu2025aiideabench}. Further, human evaluations 
on such system outputs highlight a critical gap between perceived quality and practical value \cite{si2025novelideas}. 
In scholarly research, many proposed ideas are easy to judge for novelty or clarity but difficult to assess for downstream uptake until years later. Citations are an imperfect signal, but they provide one scalable trace of how strongly a contribution is taken up by later work.These limitations motivate ideation systems that are explicitly conditioned on \emph{impact}, a delayed real-world signal for aligning models towards practical relevance and scientific quality.


While classical work on scientometrics \cite{waltman2016review, hutchins2016rcr, thelwall2013altmetrics} probes scientific impact prediction and provides a foundation for reasoning about downstream value, they fail to tie prediction as a mechanism for empowering generation systems.
Recent works extend this by forecasting high-impact topics, predicting impact of new papers, and introducing multi-dimensional benchmarks covering awards, media attention, patents, and artifact adoption \cite{gu2024forecastinghighimpactresearch, zhao2024fromwordstoworth, lu2025fromnewborntoimpact, scimpact2026benchmark, zhang2024predictingcitationimpact, zhu2024instantcitedpotential}. However, these approaches remain largely post hoc, operating on existing entities (e.g., papers or topics).
Similarly, work on future-aligned proposals and retrospective evaluation improves impact-driven ranking and judgment \cite{wang2026futurealigned, jiang2026hindsight, ajith2026prescience}, but does not optimize idea generation for expected downstream value before costly validation. Thus, current approaches focus on assessment and forecasting, not controllable, impact-aware ideation. More recently, \citet{tong2026aicanscientifictaste} train a reward model on high- vs. low-citation abstracts and use it to guide idea generation. However, their supervision remains (1) limited to binary preferences, diluting dense citation signals, and (2) conditioned on arbitrary samples rather than multiple ideas for the \emph{same} research goal, making it sensitive to surface cues (e.g. phrasing of empirical results or topic popularity cues) rather than goal-conditional impact. Moreover, optimizing responses without an explicit reward over \textit{(research goal, idea)} limits its ability to capture whether an idea is impactful for the conditioning objective. This leaves the challenge of aligning idea generation directly to impact while preserving explicit goal conditioning.

\begin{figure*}[t]
    \centering
    \includegraphics[width=17.0cm]{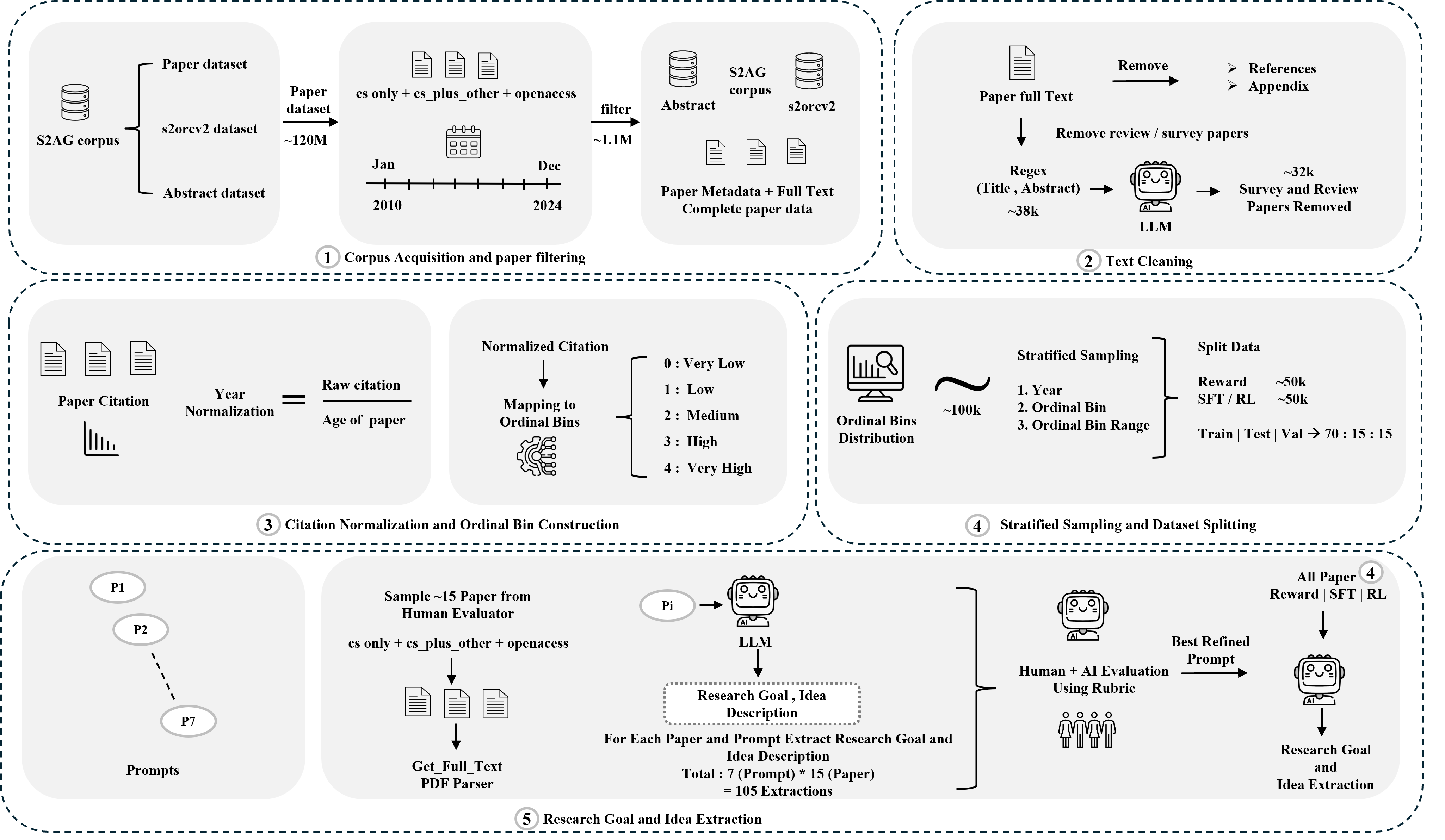}
    \caption{Dataset construction pipeline.}
    \label{fig:dataset_pipeline}
\end{figure*}


Alignment in LLMs has largely focused on generic objectives—helpfulness, harmlessness, honesty, and safety—via RLHF-, RLAIF-, and DPO-style methods \cite{wang2024alignmentSurvey, tan2025equilibrate}. We ask whether we can leverage RL via reward-models to align language models for prioritizing impactful science; a notably different challenge aimed at tying LLMs preferences towards practical, real-world guided outcomes.

In this work, we make the following contributions:
\begin{itemize}[noitemsep, topsep=2pt]
    \item We construct a large-scale dataset of research goal--idea pairs from computer science papers including interdisciplinary work, each annotated with an ordinal citation-impact label normalized by publication year.
    \item We formulate citation-aligned scientific ideation as goal-conditioned generation under a delayed uptake signal, and train a reward model over \textit{(research goal, idea)} pairs using ordinal citation labels.
    \item We align a goal-conditioned idea generator using the learned reward model, starting from supervised fine-tuning and further optimizing with reinforcement learning.
    \item We introduce a held-out, reference-grounded evaluation protocol for measuring whether generated ideas are judged impactful relative to historical ideas under the same research goal.
    \item Empirically, our reward model reaches 48.0\% accuracy on ordinal impact prediction, beating GPT-5 by over 20 points, and our RL-aligned generator improves majority-vote impact rate from 30.66\% for the base model and 26.61\% for SFT to 51.66\%.
\end{itemize}

\section{Related Work}

\paragraph{Idea Generation and Evaluation.}
A substantial body of recent work studies LLM-based scientific ideation. \textsc{SciMON}, \textsc{ResearchAgent}, \textsc{DeepInnovator}, and \textsc{IRIS} generate and refine literature-grounded ideas through novelty-oriented search, iterative revision, specialized training, and human-in-the-loop interaction \cite{wang2024scimon, baek2025researchagent, fan2026deepinnovator, garikaparthi_iris}. Complementary work has focused on evaluating idea quality through human judgments, preference modeling, or dedicated benchmarks, including \textsc{SciMuse}, IdeaBench, \textsc{AI Idea Bench 2025}, \textit{Proof of Time}, and \textit{HindSight} \cite{gu2025impact4cast, guo2024ideabench, qiu2025aiideabench, ye2026pot, jiang2026hindsight}. 

\paragraph{Citation and Scientific Impact Prediction.}
Scientific impact has traditionally been studied through citation-based indicators and their normalized variants, with complementary alternatives such as altmetrics \cite{waltman2016review, hutchins2016rcr, thelwall2013altmetrics}. More recent work extends this line by forecasting high-impact topics from evolving knowledge graphs, predicting the impact of newborn papers from textual or citation-related signals, and introducing broader benchmarks that move beyond citations to multiple dimensions of scientific influence \cite{gu2024forecastinghighimpactresearch, zhao2024fromwordstoworth, lu2025fromnewborntoimpact, scimpact2026benchmark, ajith2026prescience}. While these methods are valuable for forecasting and assessment, they largely operate over already instantiated objects such as topics, papers, or contributions, rather than directly supporting goal-conditioned idea generation.

\section{Problem Definition}

Let $\mathcal{P}=\{p_i\}_{i=1}^{N}$ denote a corpus of computer science and and related interdisciplinary papers. Each paper $p_i$ is associated with full text $x_i$ and a citation-derived impact signal. From each paper, we extract a structured pair
\begin{equation}
(g_i, d_i) = \Phi(x_i),
\end{equation}
where $\Phi$ is an LLM-based extraction function, $g_i$ denotes the \emph{research goal}, and $d_i$ denotes the corresponding \emph{idea description}. The extraction is designed to capture the core research objective and proposed idea while excluding empirical results and other post hoc evidence.

To represent impact in a form suitable for learning, each paper is assigned a year-normalized citation signal, which is further discretized into an ordinal label
$y_i \in \{0, \dots, K - 1\}$,
where the $K$ ordered categories correspond to increasing levels of scientific impact (e.g., \textit{Very Low}, \textit{Low}, \textit{Medium}, \textit{High}, and \textit{Very High}). This yields a structured dataset
\begin{equation}
\mathcal{D} = \{(g_i, d_i, y_i)\}_{i=1}^{N}.
\end{equation}

Given this dataset, our goal is to learn a conditional idea generator that, for a given research goal $g$, produces an idea description $d$ that is aligned with higher-impact ordinal labels. Formally, we seek a policy
$\pi(d \mid g)$
that generates ideas maximizing an impact-alignment objective induced by the ordinal label space (Detailed in Section \ref{sec:meth}). 
We study this problem through a three-stage framework consisting of: (i) a reward model that predicts ordinal impact from a research goal--idea pair, (ii) a supervised generator that produces idea descriptions conditioned on research goals, and (iii) a reinforcement learning stage that further aligns the generator toward higher predicted impact.

\section{Dataset Construction}

Our dataset construction pipeline consists of five steps refer figure~\ref{fig:dataset_pipeline}. 

\textbf{Step 1: Corpus Acquisition and Paper Filtering.}
We use Semantic Scholar resources—S2AG metadata \cite{wade}, the S2ORC-v2 full-text corpus, and the abstract dataset—collectively covering $\sim 120\mathrm{M}$ scholarly papers. From this pool, we select open-access papers published between January 2010 and December 2024, we choose this publication window to ensure that each paper had sufficient time to accumulate citation data by the citation collection date (10 March 2026). Papers published very recently often have artificially low citation counts due to limited exposure time rather than low scientific impact, which could incorrectly place them into lower-impact citation bins.
We then filter paper that are tagged with `Computer Science' as a field of study, including interdisciplinary works spanning computer science and other domains. This filtering yields 1.1M papers. We then align corpus identifiers across metadata, full-text, and abstract resources to construct a unified dataset containing full texts, abstracts, and metadata such as citation counts, publication dates, and field labels.

\textbf{Step 2: Text Cleaning.}
For each paper in the filtered corpus, we perform text cleaning to retain only ideation-relevant content by removing references and appendices from the full text. We also exclude review and survey papers, which primarily summarize prior work rather than present specific research goal–idea pairs. To identify these, we first apply a rule-based filter over titles and abstracts using keywords such as `review' and `survey', yielding 38K candidate papers. We then apply an LLM-based verification step to refine this set, ultimately identifying and removing 32K review or survey papers from the corpus, finally resulting into 1M papers in the cleaned corpus.

\textbf{Step 3: Citation Normalization and Ordinal Label Construction.}
To derive an impact proxy, we use citation counts scraped on March 10, 2026. Since raw citation counts are strongly influenced by publication age, we normalize citations by the age of the paper in years. Concretely, for a paper with raw citation count $c$ and age $a$, we define its citation-normalized impact score as
$x = \frac{c}{a}$,
where $x$ denotes citations per year. We then converted this continuous signals into ordinal impact bins using fixed thresholds. The thresholds are chosen to balance three considerations: preserving a meaningful ordering over citation-normalized impact, separating qualitatively different citation regimes, and maintaining sufficient support in each class for stable learning. In particular, the lowest bin captures papers with negligible yearly citation uptake, the intermediate bins capture progressively stronger but more common levels of influence, and the highest bin isolates papers with clearly exceptional citation-normalized impact. This discretization provides a robust ordinal supervision signal while reducing sensitivity to noise and heavy-tailed variation in raw citation counts \cite{bornmann2012analysepercentileimpactdata} .
\begin{align}
\textit{Zero / Very Low}   &: \; 0 \leq x \leq 0.3223, \\
\textit{Low}        &: \; 0.3223 < x < 5.0, \\
\textit{Medium}     &: \; 5.0 \leq x < 15.0, \\
\textit{High}       &: \; 15.0 \leq x < 40.0, \\
\textit{Very High}  &: \; x \geq 40.0.
\end{align}
These ordinal bins form the impact labels used in downstream reward modeling and impact-aware idea generation.\\
\textbf{Step 4: Stratified Sampling and Data Splitting.}
From the cleaned corpus, we construct a stratified sample  of 100K papers to ensure balanced coverage across both publication years and ordinal impact labels, further forming two subsets, each of 50K papers. retaining the uniform distributions.  
Each 50K sampled subset is then split into training, validation, and test partitions using a 70:15:15 ratio. One of the subset is used for reward modeling and the other is used for supervised fine-tuning (SFT), and reinforcement learning (RL). This design to have two different subset is to minimize the possibility of reward hacking arising from memorization or distribution overlap between reward modeling and policy optimization stages. 

\begin{table}[t]
\centering
\scriptsize
\setlength{\tabcolsep}{4pt}
\begin{tabular}{llcccccc}
\toprule
\textbf{Dataset} & \textbf{Split} & \textbf{Zero/VL} & \textbf{Low} & \textbf{Med.} & \textbf{High} & \textbf{V.High} & \textbf{Total} \\
\midrule
Reward & Train & 7,231 & 7,232 & 7,241 & 7,236 & 6,033 & 34,973 \\
Reward & Val   & 1,552 & 1,550 & 1,549 & 1,552 & 1,292 & 7,495 \\
Reward & Test  & 1,552 & 1,552 & 1,548 & 1,551 & 1,293 & 7,496 \\
\midrule
SFT/RL & Train & 7,239 & 7,238 & 7,236 & 7,239 & 6,038 & 34,990 \\
SFT/RL & Val   & 1,552 & 1,552 & 1,552 & 1,551 & 1,292 & 7,499 \\
SFT/RL & Test  & 1,549 & 1,552 & 1,551 & 1,550 & 1,295 & 7,497 \\
\bottomrule
\end{tabular}
\vspace{0.2cm}
\caption{Paper distribution across ordinal classes for distinct dataset subsets and splits. Counts remain approximately uniform across train, validation, and test sets, indicating successful stratification.}
\label{tab:split_ordinal}
\end{table}
\textbf{Step 5: Research Goal and Idea Extraction.}
The final stage converts each paper into a structured \textit{research goal–idea description} pair. To enable this, we first optimize the extraction prompt by designing seven candidate variants and evaluating them on 15 open-access computer science papers curated by human evaluators. For each paper, we generate outputs from all seven prompts, yielding 105 candidate extractions. These are evaluated through both human assessment and an LLM-based judge(prompt provided in Appendix \ref{lst:Prompt_evalualtion}) using a fixed rubric  
where the judge compared the extracted research goal and idea description against the full paper text. Based on this combined evaluation, we select the highest-quality prompt (prompt provided in Appendix \ref{lst:research_goal_idea_prompt}) and apply it to the sampled corpus to extract structured research goals and idea descriptions for all papers. The final dataset thus consists of triples of the form (\textit{research goal}, \textit{idea description}, \textit{ordinal impact label}). Final dataset statistics are provided in Table \ref{tab:split_ordinal}.

\begin{figure*}[t]
    \centering
    \includegraphics[width=17.0cm] {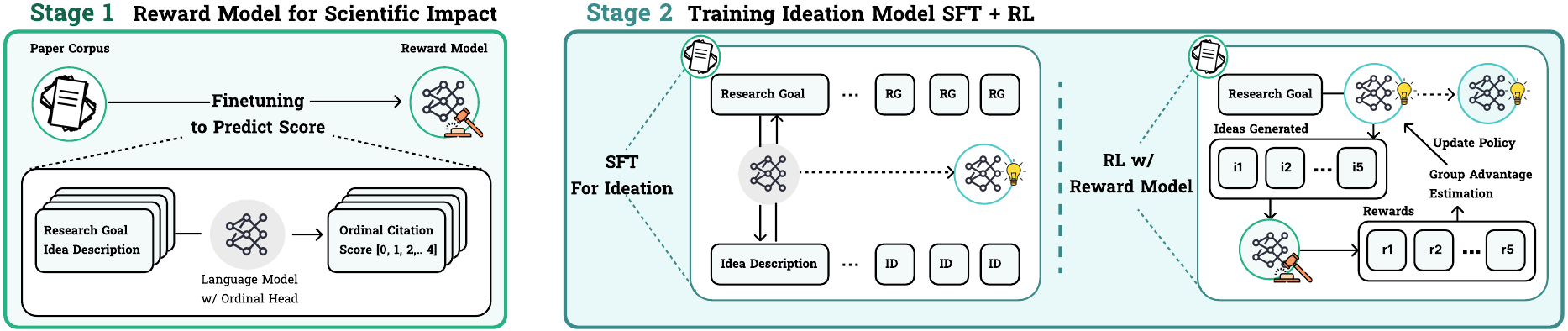}
    \caption{Impact Aligned Ideation Training formulation 
    }
    \label{fig:pipeline}
\end{figure*}

\section{Methodology}\label{sec:meth}

Our framework consists of three stages: reward modeling, supervised idea generation (SFT), and impact-aligned reinforcement learning (RL) refer figure~\ref{fig:pipeline}.

\paragraph{Reward Modeling.}
We first learn a reward model that predicts the ordinal impact label from a research goal--idea pair. Formally, let
\begin{equation}
RM_{\theta}(g,d) \rightarrow p_{\theta}(y \mid g,d),
\end{equation}
where $p_{\theta}(y \mid g,d)$ denotes an ordinal distribution over the $K$ impact levels, and $\theta$ are the reward model parameters. We implement $R_{\theta}$ by attaching an ordinal classification head based on CORN \cite{Shi_2023} to a language model backbone.

Given the training set
$\mathcal{D} = \{(g_i,d_i,y_i)\}_{i=1}^{N}$,
the reward model is trained to minimize the ordinal classification loss
\begin{equation}
\mathcal{L}_{\mathrm{RM}}(\theta)
=
\sum_{i=1}^{N}
\ell_{\mathrm{CORN}}\!\left(p_{\theta}(\cdot \mid g_i,d_i), y_i\right),
\end{equation}
where $\ell_{\mathrm{CORN}}$ denotes the CORN loss for ordinal classification \cite{Shi_2023}.

To use the reward model for policy optimization, we convert the predicted ordinal citation-impact category into a scalar reward. 

During evaluation, the reward model first predicts an ordinal citation-impact label
\begin{equation}
\hat{y} = RM_{\theta}(g,d),
\end{equation}
where $\hat{y} \in \{0,\dots,K-1\}$. The scalar reward is then obtained using a linear reward-scaling function:
\begin{equation}
r(g,d)
=
\rho(\hat{y})
=
\frac{\hat{y}}{K-1}\cdot r_{\max},
\end{equation}
where $K$ denotes the number of ordinal citation-impact categories , $r_{\max}$ is the maximum reward scaling constant and  $\rho(\cdot)$ denotes a linear reward-scaling function that maps the predicted ordinal label $\hat{y}$ to a scalar reward value.
In our implementation, we use $K=5$ and $r_{\max}=5$, resulting in scalar rewards in the range $[0,5]$. 

\paragraph{Supervised Idea Generation.}
We train a conditional generator, which is a pre-trained model $\pi_{\phi}$ to produce an idea description from a research goal $\pi_{\phi}(d \mid g)$, 
where $\phi$ are the generator parameters. The supervised fine-tuning (SFT) objective maximizes the conditional likelihood of the reference idea description given the research goal:
\begin{equation}
\mathcal{L}_{\mathrm{SFT}}(\phi)
=
- \sum_{(g_i,d_i)\in\mathcal{D}} \log \pi_{\phi}(d_i \mid g_i).
\end{equation}

This stage provides an initialization policy that can generate coherent and goal-conditioned idea descriptions before reinforcement learning, This stage provides an initialization policy that can generate coherent and goal-conditioned idea descriptions before reinforcement learning, thereby mitigating the cold-start problem commonly observed when RL optimization is initialized from weak policies \cite{ouyang2022traininglanguagemodelsfollow}.

\begin{table}[t]
\centering
\scriptsize
\setlength{\tabcolsep}{1.5pt}

\begin{tabular}{lccccc}
\toprule
\textbf{Model}
& \textbf{Acc}
& \textbf{MAE}
& \textbf{$\pm1$ Acc}
& \textbf{Spear}
& \textbf{QWK} \\
\midrule

\multicolumn{6}{c}{\textbf{Zero-shot}} \\
\midrule

Qwen3-8B
& 20.7
& 1.410
& 58.7
& 0.065
& 0.004 \\

GPT-4o
& 23.9
& 1.155
& 68.1
& 0.354
& 0.235 \\

GPT-5
& 27.7
& 1.027
& 75.3
& 0.402
& 0.338 \\

\midrule
\multicolumn{6}{c}{\textbf{Fine-tuned}} \\
\midrule

RM
& \textbf{48.0} {\color{darkgreen}(+20.3)}
& \textbf{0.632} {\color{darkgreen}(-0.39)}
& \textbf{90.6} {\color{darkgreen}(+15.3)}
& \textbf{0.756} {\color{darkgreen}(+0.35)}
& \textbf{0.754} {\color{darkgreen}(+0.41)} \\

\bottomrule
\end{tabular}

\vspace{0.2cm}

\caption{
Trained Reward Model (RM: Qwen3-8B + LoRA + CORN)  Performance comparison with zero-shot  models on the ordinal classification task.
{\color{darkgreen}Improvement of RM over GPT-5.} Acc: Accuracy, Spear: Spearman Correlation 
}
\label{tab:reward_model_results}
\end{table}

\begin{table}[t]
\centering
\scriptsize
\setlength{\tabcolsep}{10pt}

\begin{tabular}{lccccc}
\toprule
Model 
& V.Low 
& Low 
& Med. 
& High 
& V.High \\
\midrule

GPT-4o 
& 1.95 
& 1.41 
& 1.07 
& 0.96 
& \underline{\textbf{0.38}} \\

GPT-5  
& \textbf{0.66} 
& \textbf{0.86} 
& \textbf{0.94} 
& 1.02 
& 1.79 \\

Qwen3-8B 
& 3.01 
& 2.01 
& 1.00 
& \underline{\textbf{0.00}}
& 1.00 \\

RM
& \underline{\textbf{0.50}}
& \underline{\textbf{0.71}}
& \underline{\textbf{0.77}}
& \textbf{0.74} 
& \textbf{0.43} \\

\bottomrule
\end{tabular}

\vspace{0.15cm}

\vspace{0.1cm}

\caption{Class-wise Mean Absolute Error (MAE) for RM and baseline models. \underline{\textbf{Best}}, \textbf{Second Best} Performance }

\label{tab:class_mae}
\end{table}

\paragraph{Impact-Aligned Reinforcement Learning.}
Finally, we initialize a policy model with the SFT parameters and further optimize it using reinforcement learning. Let $\pi_{\psi}$ denote the policy, initialized from $\phi$. For a given research goal $g$, the policy samples a group of candidate ideas:
\begin{equation}
d^{(1)}, d^{(2)}, \dots, d^{(G)} \sim \pi_{\psi}(\cdot \mid g),
\end{equation}
where $G$ is the group size. Each generated candidate is paired with the same research goal and evaluated by the reward model:
$r^{(j)} = r(g, d^{(j)})$.
Following the Group-Relative Policy Optimization  (GRPO) \cite{shao2024deepseekmathpushinglimitsmathematical} setting, rewards within a sampled group are normalized to obtain relative advantages:
$A^{(j)} = (r^{(j)} - \mu_r)/\sigma_r$,
where $\mu_r$ and $\sigma_r$ are the mean and standard deviation of the rewards within the group. The policy is then updated to increase the likelihood of candidates with higher relative advantage. In abstract form, the optimization objective can be written as
\begin{equation}
\max_{\psi}
\;
\mathbb{E}_{g \sim \mathcal{D}, \, d^{(j)} \sim \pi_{\psi}(\cdot \mid g)}
\left[
\frac{1}{G}
\sum_{j=1}^{G}
A^{(j)} \log \pi_{\psi}(d^{(j)} \mid g)
\right].
\end{equation}

Thus, the policy is encouraged to generate idea descriptions that receive higher reward 
, i.e., generated ideas are aligned for stronger scientific impact.

\section{Experimentation and Results}

\begin{figure*}[!t]
    \centering

    \begin{subfigure}[t]{0.45\textwidth}
        \centering
        \includegraphics[width=\linewidth]{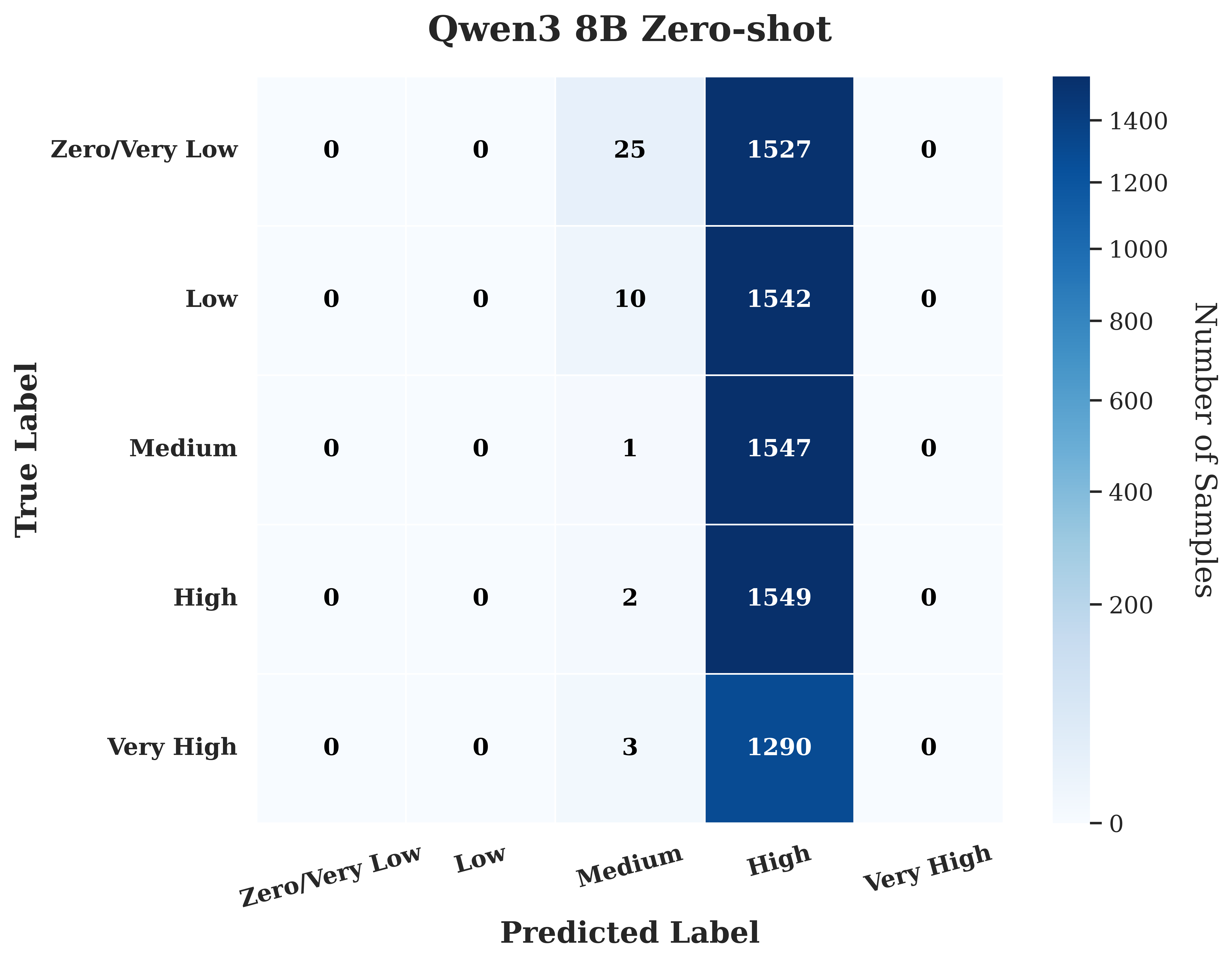}
        \caption{Qwen3-8B (Zero-shot)}
        \label{fig:qwen_zero}
    \end{subfigure}
    \hfill
    \begin{subfigure}[t]{0.45\textwidth}
        \centering
        \includegraphics[width=\linewidth]{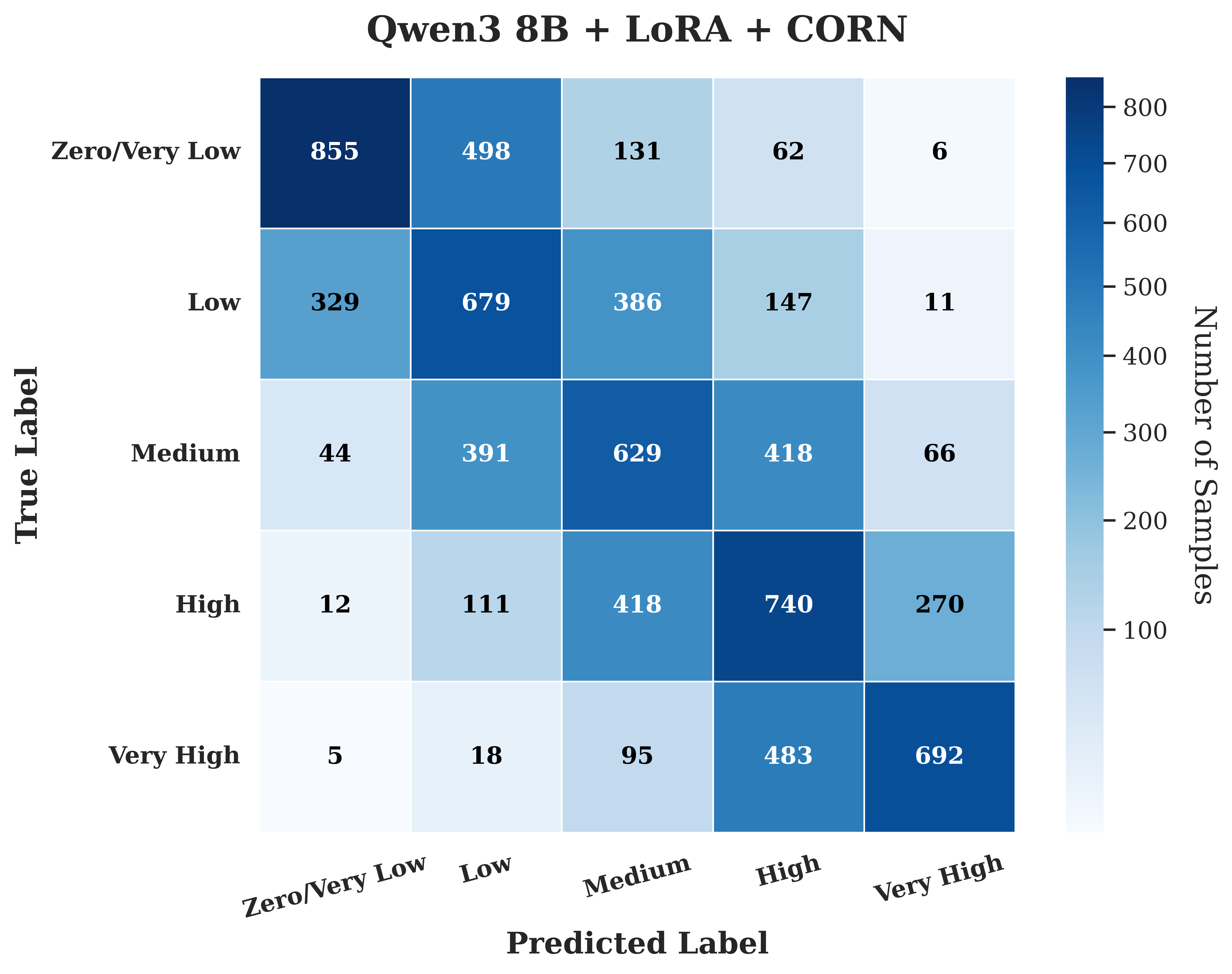}
        \caption{RM (Qwen3-8B + LoRA + CORN)}
        \label{fig:qwen_finetuned}
    \end{subfigure}

    \vspace{0.3cm}

    \caption{
    Confusion matrices 
    on the ordinal classification benchmark.
    Darker cells indicate larger sample counts.
    }

    \label{fig:confusion_matrices}
\end{figure*}

\subsection{Models and Training Details}
For reward modeling, we use Qwen3-8B based backbones, which is trained with a CORN ordinal regression head, using LoRA \cite{hu2021loralowrankadaptationlarge}   with rank $r=64$, scaling factor $\alpha=128$, dropout $0.1$, maximum sequence length 4096, and train for 3 epochs using learning rate $2\times10^{-5}$.
For supervised fine-tuning (SFT), we use Qwen3-8B with LoRA  rank $r=64$, scaling factor $\alpha=128$, and dropout $0.1$, trained for 3 epochs with maximum sequence length 2048, learning rate $2\times10^{-5}$, cosine learning-rate scheduling, and weight decay $0.01$. 
For reinforcement learning, we initialize the policy from the SFT model and optimize it using the DAPO \cite{yu2025dapoopensourcellmreinforcement} variant of GRPO with learning rate $1\times10^{-6}$, group size 4, maximum completion length 256, and KL regularization coefficient $\beta=0.01$. 
Generation during RL training uses nucleus sampling with top-$p=0.95$ and temperature $0.9$. 
Across all stages, we employ LoRA adaptation over attention and feed-forward projection layers, together with gradient checkpointing and mixed-precision BF16/FP16 training. 
Detailed RL optimization hyperparameters are provided in Table~\ref{tab:training_eval_parameters}.

\subsection{Metric}\label{sec:metric}

\paragraph{Metrics for evaluation of the Reward Model.}
We evaluate the reward model using multiple ordinal classification metrics. 
\textbf{Accuracy (Acc)} measures exact category prediction accuracy, while \textbf{Mean Absolute Error (MAE)} measures the average ordinal distance between predicted and ground-truth citation-impact labels. 
\textbf{$\pm1$ Accuracy} considers a prediction correct if it falls within one adjacent ordinal category of the ground-truth label. 
\textbf{Spearman Correlation (Spear)} evaluates rank correlation between predicted and true citation-impact levels, and \textbf{Quadratic Weighted Kappa (QWK)} measures ordinal agreement while assigning larger penalties to predictions that are farther from the correct citation-impact category.

\paragraph{Metrics for the evaluation of the Idea Generator.}
We propose a novel metric to evaluate the impactfulness of generated research ideas: \textbf{Reference-Grounded Citation-Weighted Idea Impact Evaluation}
(\textbf{RGCW-IIE}).
The evaluation is performed by comparing each generated idea against a ground-truth reference idea under the same research goal. Unlike similarity-based
metrics, the proposed evaluation does not require the generated idea to exactly
match the reference idea. Instead, it measures whether the generated idea is
judged to be scientifically impactful relative to the ground-truth idea, and
weights this judgment using the citation-normalized impact label of the
reference idea.

Let
$\mathcal{D} = \{(g_i, d_i, y_i)\}_{i=1}^{N}$
denote the evaluation dataset, where $g_i$ is the research goal, $d_i$ is the
ground-truth reference idea description, and
$y_i \in \{0,1,\ldots,K-1\}$
is the ordinal year-normalized citation label assigned to the ground-truth idea.
The label $y_i$ represents the citation impact of the reference idea after normalizing for publication year.
Let
$\mathcal{M} \in 
\{\mathcal{M}_{\mathrm{base}},
\mathcal{M}_{\mathrm{SFT}},
\mathcal{M}_{\mathrm{RL}}\}$
denote a candidate idea-generation model, corresponding to the baseline,
supervised fine-tuned, and reinforcement-learned models, respectively. For each
research goal $g_i$, the model generates an idea (prompt shown in Appendix~\ref{lst:generation_prompt}) 
$\hat{d}_i^{\mathcal{M}} = \mathcal{M}(g_i)$.
Given the research goal $g_i$, the ground-truth idea $d_i$, and the generated
idea $\hat{d}_i^{\mathcal{M}}$, we use a strong evaluator model $\mathcal{E}$, i.e GPT-5.1, GPT-4.1, GPT-4o, to judge whether the
generated idea is impactful relative to the ground-truth idea. The evaluator
produces a binary impact judgment and a natural-language rationale explaining why the generated
idea is considered impactful or not impactful.
($z_i^{\mathcal{M}},  q_i^{\mathcal{M}}) =
\mathcal{E}(g_i, d_i, \hat{d}_i^{\mathcal{M}})
$,
where $q_i^{\mathcal{M}}$ denotes the evaluator's textual justification and
\[
z_i^{\mathcal{M}} =
\begin{cases}
1, & \text{if } \hat{d}_i^{\mathcal{M}} \text{ is judged impactful relative to } d_i, \\
0, & \text{otherwise.}
\end{cases}
\]
The evaluator is instructed to make its decision by considering the relevance of the generated idea to the research goal, based on its 
scientific contribution and potential impact. Prompt~\ref{lst:evaluator_prompt}  


To incorporate the citation impact of the reference idea into the evaluation, we
define the instance-level \textbf{Reference-Grounded Citation-Weighted Impact Score} as
$s_i^{\mathcal{M}}
=
z_i^{\mathcal{M}}(y_i + 1)$
Equivalently,
\[
s_i^{\mathcal{M}} =
\begin{cases}
y_i + 1, & \text{if } \hat{d}_i^{\mathcal{M}} \text{ is judged impactful relative to } d_i, \\
0, & \text{otherwise.}
\end{cases}
\]

The addition of one ensures that an impactful generated idea receives a positive score even when the corresponding reference idea belongs to the lowest ordinal citation-impact class. Therefore, the reward reflects both the binary impactfulness judgment and the citation-normalized importance of the reference
idea. The overall \textbf{RGCW-IIE} score of model $\mathcal{M}$ is computed as the average
score over all evaluation instances:
$=
\frac{1}{N}
\sum_{i=1}^{N}
s_i^{\mathcal{M}}$

A higher value of  \textbf{RGCW-IIE} 
indicates that the model more frequently generates ideas judged to be impactful relative to the
ground-truth reference ideas, particularly for examples whose references have higher citation-normalized impact.
For bounded comparison, we define the normalized score:
\[
\mathrm{nRGCW\text{-}IIE}(\mathcal{M})
=
\frac{
\sum_{i=1}^{N} z_i^{\mathcal{M}}(y_i+1)
}{
\sum_{i=1}^{N} (y_i+1)
}.
\]
Since $y_i \in \{0,1,\ldots,K-1\}$ and $z_i^{\mathcal{M}} \in \{0,1\}$, the normalized score satisfies
$0 \leq \mathrm{nRGCW\text{-}IIE}(\mathcal{M}) \leq 1$.

We also define \textbf{IR} (Impact Rate) which denotes the percentage of generated ideas judged as impactful by the evaluator, while \textbf{I'Idea} (Impactful Ideas) denotes the absolute number of generated ideas classified as impactful.

\subsection{Reward Model Performance}

We first evaluate the reward model on the ordinal citation-impact prediction task. As shown in Table~\ref{tab:reward_model_results}, the proposed RM substantially outperforms all zero-shot baselines across every metric. In particular, it 
exceeds the strongest zero-shot baseline GPT-5.
The same trend is also reflected in rank-sensitive and agreement-based metrics, where the fine-tuned reward model reaches a Spearman correlation of 0.756 and a quadratic weighted kappa (QWK) of 0.754. These results indicate that impact prediction benefits strongly from goal-conditioned supervised adaptation, and that the learned reward model captures ordinal structure in citation-normalized impact substantially better than prompt-only zero-shot evaluation.

Table~\ref{tab:class_mae} further shows that the fine-tuned reward model improves class-wise prediction quality across most impact bands. Relative to zero-shot baselines, the model reduces error substantially in the zero/very-low, low, medium, and high categories, while remaining competitive even on the very-high-impact category, which is typically the most difficult due to its rarity and broader semantic variability. 
Figure~\ref{fig:confusion_matrices} compares the confusion matrices of the zero-shot and fine-tuned Qwen3-8B models on the ordinal citation-impact classification benchmark. 
The zero-shot Qwen3-8B model exhibits severe prediction collapse, assigning the majority of samples to the \textit{High} citation-impact category irrespective of the ground-truth label. In contrast, the  fine-tuned model with the CORN ordinal classification head demonstrates substantially improved ordinal discrimination, with predictions concentrated more closely along the diagonal entries of the confusion matrix, indicating stronger alignment between predicted and ground-truth citation-impact categories.

\begin{table}[t]
\centering
\scriptsize
\setlength{\tabcolsep}{8pt}
\renewcommand{\arraystretch}{1.1}

\begin{tabular}{llccccc}
\toprule
Eval. 
& Method 
& IR 
& I'Ideas 
& RGCW 
& nRGCW \\
\midrule

\multirow{3}{*}{GPT-4o}
& Base  & 32.66 & 2440 & 0.432 & 0.147 \\
& SFT   & 27.30 & 2040 & 0.320 & 0.109 \\
& RL    & \textbf{50.58} & \textbf{3779} & \textbf{0.709} & \textbf{0.242} \\
\midrule

\multirow{3}{*}{GPT-4.1}
& Base  & 52.60 & 3930 & 0.965 & 0.329 \\
& SFT   & 59.53 & 4448 & 1.171 & 0.399 \\
& RL    & \textbf{81.80} & \textbf{6112} & \textbf{1.753} & \textbf{0.597} \\
\midrule

\multirow{3}{*}{GPT-5.1}
& Base  & 3.45 & 258 & 0.037 & 0.013 \\
& SFT   & 4.80 & 359 & 0.060 & 0.021 \\
& RL    & \textbf{19.90} & \textbf{1487} & \textbf{0.275} & \textbf{0.094} \\
\midrule

\multirow{3}{*}{Maj. Vote}
& Base  & 30.66 & 2291 & 0.398 & 0.136 \\
& SFT   & 26.61 & 1988 & 0.314 & 0.107 \\
& RL    & \textbf{51.66} & \textbf{3860} & \textbf{0.763} & \textbf{0.260} \\
\bottomrule
\end{tabular}
\vspace{0.1cm}

\caption{
Reference-grounded citation-weighted idea impact evaluation results. 
IR: Impact Rate, I'Ideas: Impactful Ideas RGCW: Reference-Grounded Citation-Weighted score, nRGCW: Normalized Reference-Grounded Citation-Weighted score, Eval.: Evaluator, Maj. Vote: Majority Vote. Number of evaluation test samples ($N$): 7472. 
The upper bound of RGCW is approximately $2.93$ based on ground-truth citation-impact labels.
}
\label{tab:impact_eval_all_evaluators}
\end{table}


\subsection{Impact-Aligned Idea Generation}

We next evaluate whether optimizing against the learned reward improves the impactfulness of generated research ideas. Table~\ref{tab:impact_eval_all_evaluators} reports results under the proposed Reference-Grounded Citation-Weighted Idea Impact Evaluation (RGCW-IIE) protocol (Discussed in Section \ref{sec:metric}) across three evaluator models and a majority-vote aggregation. The overall pattern is highly consistent: the RL-aligned model outperforms both the baseline and the supervised fine-tuned model under every evaluator across all metrics.
In particular, GPT-4.1, which tends to assign comparatively higher impact scores, still preserves the same ranking across all three methods. In contrast, GPT-5.1 is considerably more conservative in its absolute judgments, yet it assigns the highest score to the RL model by a clear margin.

These results suggest that reward-guided policy optimization does more than simply preserve fluency or goal relevance: it biases generation toward ideas that are more likely to be judged as impactful relative to historically grounded references. Importantly, the RL model improves over SFT even though both models are conditioned on the same research goal and use the same prompt format. This indicates that supervised training alone is insufficient to align idea generation with downstream impact objectives, whereas reinforcement learning with an explicit impact-sensitive reward provides a more effective mechanism for steering the model toward practically stronger research directions.

\begin{table}[h]
\centering
\scriptsize
\setlength{\tabcolsep}{8pt}
\renewcommand{\arraystretch}{1.1}

\begin{tabular}{lcccccc}
\toprule
Pattern
& \multicolumn{2}{c}{GPT-5.1}
& \multicolumn{2}{c}{GPT-4.1}
& \multicolumn{2}{c}{GPT-4o} \\
\cmidrule(lr){2-3}
\cmidrule(lr){4-5}
\cmidrule(lr){6-7}

& Cnt & \%
& Cnt & \%
& Cnt & \% \\
\midrule

RL only
& 1252 & 76.3
& 1130 & 30.2
& 1206 & 37.6 \\

RL+SFT
& 149 & 9.1
& 1427 & 38.1
& 653 & 20.4 \\

Base+RL
& 74 & 4.5
& 821 & 21.9
& 829 & 25.9 \\

Base only
& 156 & 9.5
& 251 & 6.7
& 397 & 12.4 \\

Base+SFT
& 11 & 0.7
& 116 & 3.1
& 119 & 3.7 \\

\midrule
Total
& \textbf{1642} & \textbf{100}
& \textbf{3745} & \textbf{100}
& \textbf{3204} & \textbf{100} \\

\bottomrule
\end{tabular}

\vspace{0.1cm}

\caption{
Distribution of evaluator-confirmed impact patterns across generation methods. 
Cnt: Count, Base: Baseline.
Percentages are normalized within each evaluator.
}
\label{tab:impact_pattern_distribution}
\end{table}
Table~\ref{tab:impact_pattern_distribution} reports the distribution of \emph{impact permutations} across generation methods. Here, a permutation denotes the subset of methods whose generated ideas are judged impactful for the same evaluation sample. For example, the \emph{RL only} permutation counts samples for which only the RL-generated idea is judged impactful, while both the baseline and SFT generations are not. Likewise, \emph{RL + SFT only} denotes samples where both RL and SFT are judged impactful but the baseline is not. The results show that impactful outcomes are disproportionately concentrated in permutations involving the RL model. 

\subsection{Analysis}

A qualitative reading of generated outputs reveals a consistent difference in the type of reasoning encouraged by each training stage. Baseline generations tend to remain abstract, generic, or loosely specified, often describing high-level mechanisms without enough operational detail to support implementation or evaluation. SFT improves structure and topical grounding, but many outputs still remain moderately generic, frequently restating the goal in cleaner prose without adding sufficiently concrete execution logic. In contrast, the RL-aligned model more often produces implementation-oriented ideas with clearer module interactions, decision pathways, operational constraints, and engineering specificity.  Sample generated outputs with Qualitative Comparison in Tables \ref{tab:qualitative_comparison_vit_fas} , \ref{tab:qualitative_comparison_emergency_llm} , \ref{tab:qualitative_comparison_ssl_fewshot} , \ref{tab:qualitative_comparison_3} and \ref{tab:qualitative_comparison_4} in the appendix.
This qualitative difference aligns with the quantitative gains in RGCW-IIE. 

\section{Conclusion}

We present an impact-aligned framework for scientific idea generation that uses citation-normalized downstream influence as a scalable world-feedback signal. Starting from a large corpus of scientific papers, we constructed a dataset of research goal--idea pairs with ordinal impact labels, trained a goal-conditioned reward model to estimate expected impact, and used this reward to align an idea generator through reinforcement learning. Across both reward-model evaluation and downstream idea-generation evaluation, the proposed framework consistently outperforms zero-shot and supervised baselines.

Our results suggest that scientific impact, despite being delayed and noisy, can still provide a practical alignment target for open-ended generation. Rather than optimizing only for surface-level properties such as novelty or plausibility, the proposed method encourages models to generate ideas that are more likely to resemble historically impactful scientific directions. More broadly, this work highlights the potential of moving beyond generic alignment objectives toward outcome-grounded objectives that better reflect the long-term value of scientific work.

Future work can explore richer notions of scientific impact beyond citation counts, including reproducibility, software adoption, clinical relevance, and long-term societal influence. Improving the quality of automatic goal--idea extraction and extending the framework to broader scientific domains are also important directions. Additionally, interactive human--AI ideation and stronger reasoning capabilities may further improve the quality, feasibility, and usefulness of generated scientific ideas.
\begin{table*}[t]
\centering

\begin{tcolorbox}[
enhanced,
width=\textwidth,
colback=white,
colframe=black,
fonttitle=\bfseries,
sharp corners,
boxrule=0.6pt,
left=1.5pt,
right=1.5pt,
top=1.5pt,
bottom=1.5pt
]

\fontsize{8.3pt}{9.2pt}\selectfont
\setlength{\tabcolsep}{2.8pt}
\renewcommand{\arraystretch}{0.94}

\textbf{Research Goal:}

\vspace{1pt}
The core objective of this research is to investigate the fundamental natures of vision transformer (ViT) in multimodal face anti-spoofing (FAS), emphasizing modality-aware inputs, suitable multimodal pre-training, and efficient finetuning. This research addresses the challenge of enhancing FAS systems' robustness against various presentation attacks using multimodal inputs, specifically RGB, Infrared (IR), and Depth modalities. The expected outcome is to develop models that efficiently process these inputs to improve FAS performance without relying on extensive labeled datasets. This study aims to bridge the existing gaps in ViT’s application within multimodal FAS by providing insights and solutions for better feature utilization and model training.

\vspace{0.15cm}
\textbf{Color Legend:}

\vspace{1pt}
\baselinehl{Baseline} weak/abstract technical grounding \hspace{0.22cm}
\sfthl{SFT} moderate structured grounding \hspace{0.22cm}
\rlhl{RL} implementation-oriented grounding

\vspace{0.15cm}

\begin{tabularx}{\textwidth}{
>{\raggedright\arraybackslash}p{1.3cm}
>{\raggedright\arraybackslash}X
}
\toprule
\textbf{Method} & \textbf{Generated Idea Description} \\
\midrule

\textbf{Baseline} &
This research explores the application of vision transformers (ViT) in multimodal face anti-spoofing by designing a framework that processes RGB, infrared (IR), and depth modalities through modality-aware input encodings. Each modality is independently transformed into patch embeddings via separate ViT encoders, which capture spatial hierarchies specific to their respective input types. \baselinehl{These embeddings are then concatenated and passed through a fusion module that learns cross-modal interactions while preserving modality-specific features.} A multimodal pre-training strategy is employed, where the model is trained on a large-scale dataset with unlabeled samples to align features across modalities and enhance generalization. During fine-tuning, task-specific losses are applied to refine the model's ability to distinguish between genuine and spoofed faces based on multimodal cues. \baselinehl{The training procedure involves alternating between pre-training objectives and task-specific supervision to balance representation learning with downstream performance.} The inference phase utilizes the pre-trained and fine-tuned model to extract features from unseen test samples, which are then classified using a final projection layer. Attention mechanisms within the ViT architecture are leveraged to dynamically weigh the importance of different modalities during decision-making. \baselinehl{The design ensures efficient computation by minimizing redundant processing while maintaining discriminative power across modalities.} The framework emphasizes the integration of modality-specific knowledge with global contextual understanding to achieve robust feature representation. By structuring the information flow to accommodate diverse input types, the model enables effective adaptation to varying spoofing techniques. The overall approach aims to establish a principled way of utilizing ViT in multimodal settings without requiring extensive labeled data for training. \\
\midrule

\textbf{SFT} &
This work introduces a methodological framework centered around three key aspects of multimodal FAS using vision transformer architectures. Initially, the paper highlights the importance of modality-aware designs, proposing an enhanced architecture for ViT that integrates both IR and RGB modalities through a multi-branch structure with dedicated patch embedding blocks. \sfthl{This design allows for the simultaneous processing of features from different modalities while preserving their distinct characteristics.} Following this, it advocates for a fine-tune-friendly pre-training strategy called multimodal contrastive learning (MCL), which leverages unlabelled multimodal data to create a robust feature space that facilitates effective knowledge transfer to downstream FAS tasks. \sfthl{Lastly, the methodology suggests that fine-tuning should be performed on both modalities during training to maximize feature utilization, contrasting previous approaches that emphasized singular modality focus.} The training involves optimizing a contrastive loss function derived from MCL, ensuring that all patches across modalities are treated equally, thus addressing the inherent asymmetry often found in conventional single-modality models. \sfthl{By implementing these strategies, the proposed framework demonstrates a comprehensive enhancement in the multimodal recognition capabilities required for face anti-spoofing applications.} \\
\midrule

\textbf{RL} &
This work introduces Modality-Aware Vision Transformer (MAViT), a novel architecture designed to leverage multimodal features effectively for face anti-spoofing. MAViT consists of a modular design that includes a Multi-modal Input Processor (MIP), which extracts diverse feature sets—such as pixel-level, texture-level, and geometric-level features—from three different modalities: RGB, IR, and Depth. \rlhl{These modal-specific features are processed by independent vision transformers to form multi-layer feature representations, which are then integrated into a Multi-Modal Fusion Block (MMFB) to learn cross-modality representations.} The MMFB utilizes a hierarchical fusion mechanism that aggregates information at both feature and token levels, allowing it to capture complementary characteristics while retaining modality-specific attributes. \rlhl{To enhance the robustness of MAViT, the paper also presents a specialized Pre-training Task (APT) that employs two self-supervised tasks: a Contrastive Task (CT) focusing on cross-modal contrastive learning between real and fake samples, and a Rotation Task (RT) that encourages learning of discriminative cross-modal representation through joint rotation operations on the multi-modal inputs.} This methodology allows MAViT to achieve state-of-the-art performance on two benchmark datasets with significantly fewer labeled samples than traditional methods. \rlhl{Furthermore, a two-stage finetuning strategy is proposed to optimize MAViT's efficiency, where pre-finetuning occurs with minimal labeled data to stabilize performance, followed by fine-tuning to enhance accuracy further.} The overall training process ensures that the model learns to exploit rich, modality-aware features effectively, ultimately achieving superior generalization capabilities for real-world FAS applications. \\
\bottomrule
\end{tabularx}

\vspace{0.15cm}

\noindent\textbf{Observation:}
The RL-generated idea is identified as impactful because it proposes a concrete, technically detailed ViT-based multimodal FAS framework that directly targets modality-aware inputs, multimodal pre-training, and efficient finetuning. Compared to the Baseline idea, which mainly describes a standard multimodal ViT pipeline using separate encoders, feature concatenation, fusion, unlabeled pretraining, and attention-based modality weighting, the RL approach introduces a more specialized and structured architecture. Compared to the SFT idea, which relies on a multi-branch ViT design and generic multimodal contrastive learning, the RL approach provides more clearly defined FAS-specific mechanisms. It combines a Multi-modal Input Processor, independent modality-specific ViTs, a hierarchical Multi-Modal Fusion Block, tailored self-supervised pre-training tasks, and a two-stage finetuning strategy. The contrastive and rotation-based pre-training tasks are designed to improve cross-modal real-versus-fake representation learning, while hierarchical feature- and token-level fusion helps capture complementary modality information. These components make the RL idea more specific, innovative, and aligned with the research goal than the Baseline and SFT ideas, which are judged less impactful because they are more routine, generic, and under-specified.

\end{tcolorbox}

\vspace{2pt}
\caption{Qualitative Comparison of Baseline, SFT, and RL Generated Research Ideas}
\label{tab:qualitative_comparison_vit_fas}

\end{table*}

\section{Limitations}

Firstly, citation count is an imperfect proxy for scientific impact: it is influenced by field norms, publication age, visibility, and external social factors, and therefore does not fully capture the true value of a research idea. 

Second, our dataset construction pipeline depends on automatic extraction of research goals and ideas from papers. While this enables scalable data collection, it may omit nuance, compress technical detail, or propagate extraction errors into both the reward and generator model. 

Third, our evaluation protocol relies on LLM-based comparative judgments, which may themselves introduce stylistic or domain-specific biases. While we reduce this risk through providing ground truth idea in the reference, multiple evaluators and majority-vote aggregation, the evaluation remains only an approximation of real scientific usefulness.

Finally, our experiments are restricted to the computer science and its related domain and impact signals derived from published literature. 

\clearpage

\section*{Impact Statement}
Our work explore the use of citation-normalized impact signals as outcome grounded feedback for aligning large language models (LLMs) in scientific ideation. By adding this delayed signals of research outcomes into model training, our approach aims to steer generative systems towards ideas with higher expected scientific impact. Such system have potential to support researchers in identifying promising research directions, accelerating discovery, and also improving efficiency of scientific exploration.
\par With this there arises several ethical and societal consideration.First citation-based impact signals is an imperfect and historically contingent proxy for scientific value. Citation may reflect existing community bias, popularity effects, institutional advantage, or dominant research paradigms.Second, deploying LLM-based ideation systems in scientific workflows raises concerns about authorship, intellectual ownership, and attribution of credit.
\par At the same time, impact-aware ideation models may broaden access to high level research support by lowering barriers to brainstorming and proposal development, particularly for early-career researcher or those in under-resourced institutions. By framing impact as a measurable but imperfect feedback signals, this work contributes to broader discussion on outcome-based alignment in generative systems.
Overall this paper advances methods for aligning LLMs using real-world outcome signals. With appropriate safeguards and critical evaluation, impact-aware ideation models have the potential to positively contribute to research ecosystem.


\bibliography{example_paper}
\bibliographystyle{icml2026}

\clearpage
\appendix
\section{Appendix}
This appendix provides supplementary tables, examples and a note on ethical considerations.

\vspace{1cm}
\begin{table}[H]
\centering
\begin{tabular}{|l|r|}
\hline
\textbf{Item} & \textbf{Value} \\
\hline
Number of samples & 100{,}000 \\
Input tokens per sample & 8{,}000 \\
Output tokens per sample & 500 \\
Total input tokens & 800{,}000{,}000 \\
Total output tokens & 50{,}000{,}000 \\
Input price per 1M tokens & \$0.15 \\
Output price per 1M tokens & \$0.60 \\
Input cost & \$120 \\
Output cost & \$30 \\
Final cost & \$150 \\
Batch API discount & 50\% \\
\textbf{Total cost} & \textbf{\$75} \\
\hline
\end{tabular}
\vspace{0.5cm}
\caption{Estimated cost for processing 100{,}000 samples with GPT-4o-mini}
\label{tab:gpt4omini_cost}
\end{table}

Table~\ref{tab:gpt4omini_cost} presents the estimated cost of generating research idea descriptions for 100{,}000 samples using GPT-4o-mini. 
The estimation assumes an average input context length of 8{,}000 tokens and an average generated output length of 500 tokens per sample. 
Under the GPT-4o-mini pricing scheme, the total processing cost is approximately \$75 after using batch mode for extraction, demonstrating the scalability and cost-effectiveness of constructing large-scale research idea generation datasets using modern language models.

\vspace{2cm}
\begin{table}[H]
\centering

\begin{tabular}{lcccc}
\hline
\textbf{Dataset} & \textbf{Train} & \textbf{Validation} & \textbf{Test} & \textbf{Total} \\
\hline
Reward  & 34{,}973 & 7{,}495 & 7{,}496 & 49{,}964 \\
SFT/RL    & 34{,}990 & 7{,}499 & 7{,}497 & 49{,}986 \\
\hline
\end{tabular}
\vspace{0.3cm}
\caption{Dataset statistics after Batch API generation. The original target size was 50K instances per dataset; differences indicate failed or missing generations.}
\label{tab:dataset_split_stats}
\end{table}

\begin{table*}[!t]
\begin{adjustbox}{width=\textwidth}
\begin{tabular}{l l c c c c c}
\toprule
\textbf{Feature Mode} & \textbf{Model} & \textbf{Accuracy (\%) $\uparrow$} & \textbf{MAE $\downarrow$} & \textbf{$\pm1$ Acc (\%) $\uparrow$} & \textbf{Spearman $\uparrow$} &  \textbf{QWK $\uparrow$}\\
\midrule

\multicolumn{7}{c}{\textit{Goal Only}} \\
\midrule
goal-only & Logistic Regression (multiclass) & 34.46 & 1.051 & 72.71 & 0.481 & 0.481 \\
goal-only & Linear SVM & 33.43 & 1.092 & 71.28 & 0.456 & 0.456 \\
goal-only & Random Forest & 32.22 & 1.195 & 66.66 & 0.401 & 0.401 \\
goal-only & Ridge Regression & 29.55 & 0.947 & 78.42 & 0.494 & 0.446 \\
goal-only & Ordinal Logistic  & 33.84 & 1.042 & 72.63 & 0.506 & 0.503 \\

\addlinespace[4pt]
\midrule
\multicolumn{7}{c}{\textit{Idea Only}} \\
\midrule
idea-only & Logistic Regression (multiclass) & 35.59 & 1.044 & 72.27 & 0.488 & 0.488 \\
idea-only & Linear SVM & 34.53 & 1.082 & 70.74 & 0.469 & 0.468 \\
idea-only & Random Forest & 32.79 & 1.206 & 66.17 & 0.434 & 0.428 \\
idea-only & Ridge Regression & 29.74 & 0.930 & 79.63 & 0.519 & 0.468 \\
idea-only & Ordinal Logistic  & 34.69 & 1.023 & 72.83 & 0.524 & 0.521 \\

\addlinespace[4pt]
\midrule
\multicolumn{7}{c}{\textit{Goal + Idea}} \\
\midrule
goal+idea & Logistic Regression (multiclass) & 36.15 & 1.001 & 74.56 & 0.513 & 0.513 \\
goal+idea & Linear SVM & 34.42 & 1.063 & 71.92 & 0.471 & 0.470 \\
goal+idea & Random Forest & 34.00 & 1.152 & 68.29 & 0.452 & 0.450 \\
goal+idea & Ridge Regression & 32.24 & 0.902 & 80.20 & 0.541 & 0.507 \\
goal+idea & Ordinal Logistic  & 35.71 & 0.982 & 74.85 & 0.547 & 0.545 \\

\addlinespace[4pt]
\midrule
\multicolumn{7}{c}{\textit{Goal + Idea + Difference Features}} \\
\midrule
goal+idea+diff & Logistic Regression (multiclass) & 34.50 & 1.037 & 73.08 & 0.487 & 0.487 \\
goal+idea+diff & Linear SVM & 33.06 & 1.088 & 71.04 & 0.459 & 0.458 \\
goal+idea+diff & Random Forest & 33.18 & 1.166 & 68.26 & 0.443 & 0.442 \\
goal+idea+diff & Ridge Regression & 31.98 & 0.913 & 79.60 & 0.533 & 0.499 \\
goal+idea+diff & Ordinal Logistic & 34.85 & 1.004 & 73.98 & 0.533 & 0.531 \\

\addlinespace[4pt]
\midrule
\multicolumn{7}{c}{\textit{LLM Reward Model}} \\
\midrule
goal+idea & Qwen3-8B + LoRA + CORN ($r{=}64,\alpha{=}128$) & \textbf{48.00} & \textbf{0.632} & \textbf{90.60} & \textbf{0.756} & \textbf{0.754} \\

\bottomrule
\end{tabular}
\end{adjustbox}
\vspace{4pt}
\caption{Test-set performance of classical baselines under different feature settings, together with the Qwen3-8B reward model fine-tuned with LoRA and a CORN ordinal head. Accuracy and $\pm1$ Accuracy are reported in percentage. Bold indicates the best overall result in each metric column.}

\smallskip
Table~\ref{tab:test_baselines_lora} reports the performance of non-LLM TF-IDF baselines for ordinal citation-impact prediction. The baselines evaluate four feature settings: using only the research goal, using only the idea description, concatenating goal and idea features, and adding explicit difference features between the two representations. For each setting, we compare multiclass classifiers, regression-as-ordinal prediction, and ordinal regression models.
\vspace{0.2cm}
\par Overall, combining the research goal and idea description provides the strongest non-LLM performance, suggesting that both the problem context and proposed solution contain complementary signals for predicting citation-normalized impact.
\vspace{0.2cm}
\par The Qwen3-8B reward model substantially outperforms all TF-IDF baselines across all evaluation metrics, demonstrating the advantage of contextual language representations for modeling nuanced semantic relationships between research goals, proposed ideas, and expected scientific impact. Unlike sparse lexical TF-IDF features, the LLM-based reward model can capture higher-level semantic structure, reasoning patterns, and implicit scientific relevance signals that are important for ordinal citation-impact prediction.

\label{tab:test_baselines_lora}
\end{table*}

\begin{figure}[t]
    \centering
    \includegraphics[width=\linewidth]{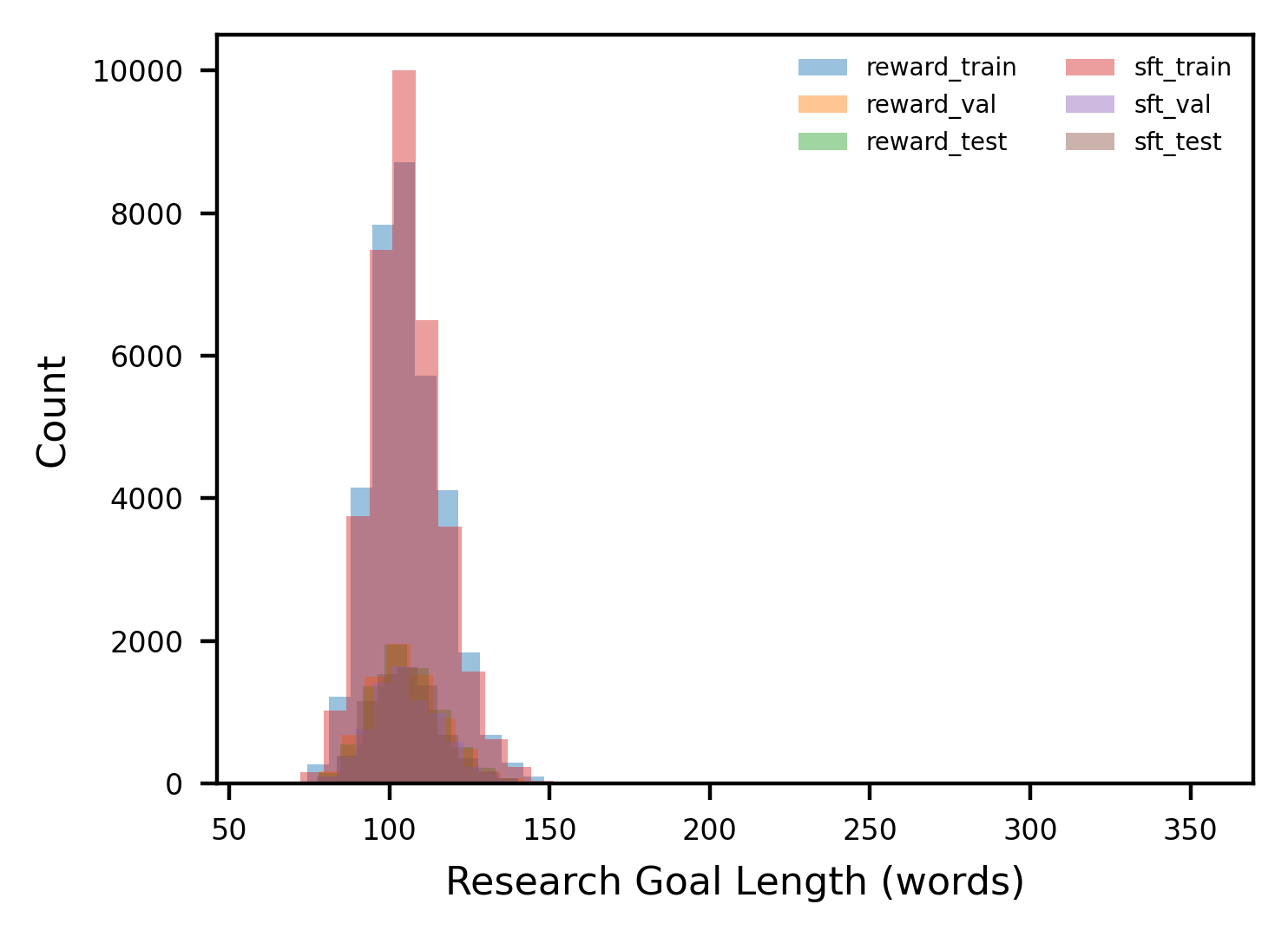}
    \caption{
    Distribution of research goal lengths across the reward and SFT dataset splits.
    The research goal lengths approximately follow a normal distribution with a peak around 100 words, indicating consistent problem-context formulation across samples.
    }
    \label{fig:goal_length_distribution}
\end{figure}

\begin{figure}[t]
    \centering
    \includegraphics[width=\linewidth]{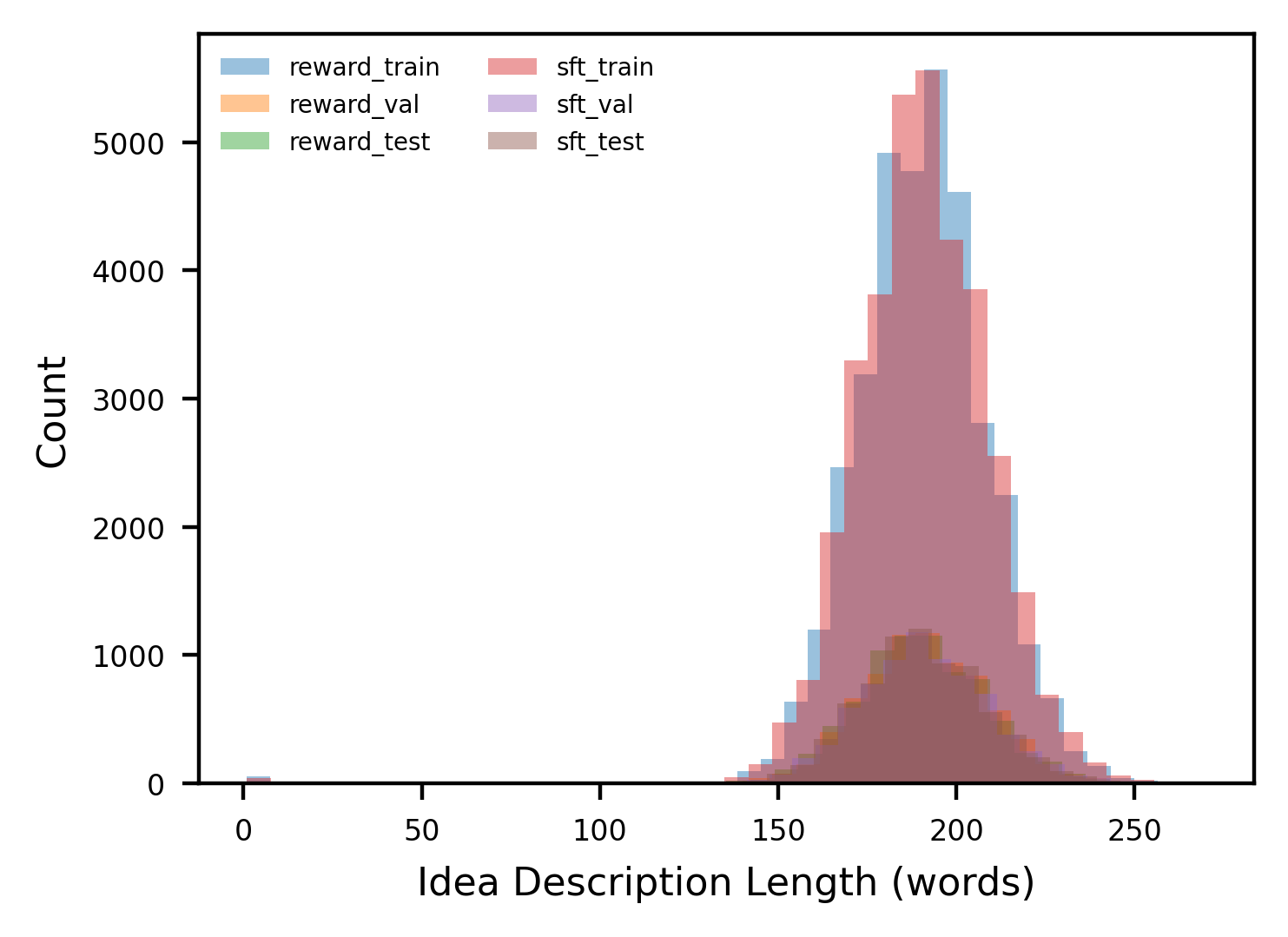}
    \caption{
    Distribution of idea description lengths across the reward and SFT dataset splits.
    The idea descriptions exhibit a normal distribution with a peak around 200 words, reflecting richer methodological and implementation-oriented detail in the proposed research ideas.
    }
    \label{fig:idea_length_distribution}
\end{figure}

\begin{figure}[t]
    \centering
    \includegraphics[width=\linewidth]{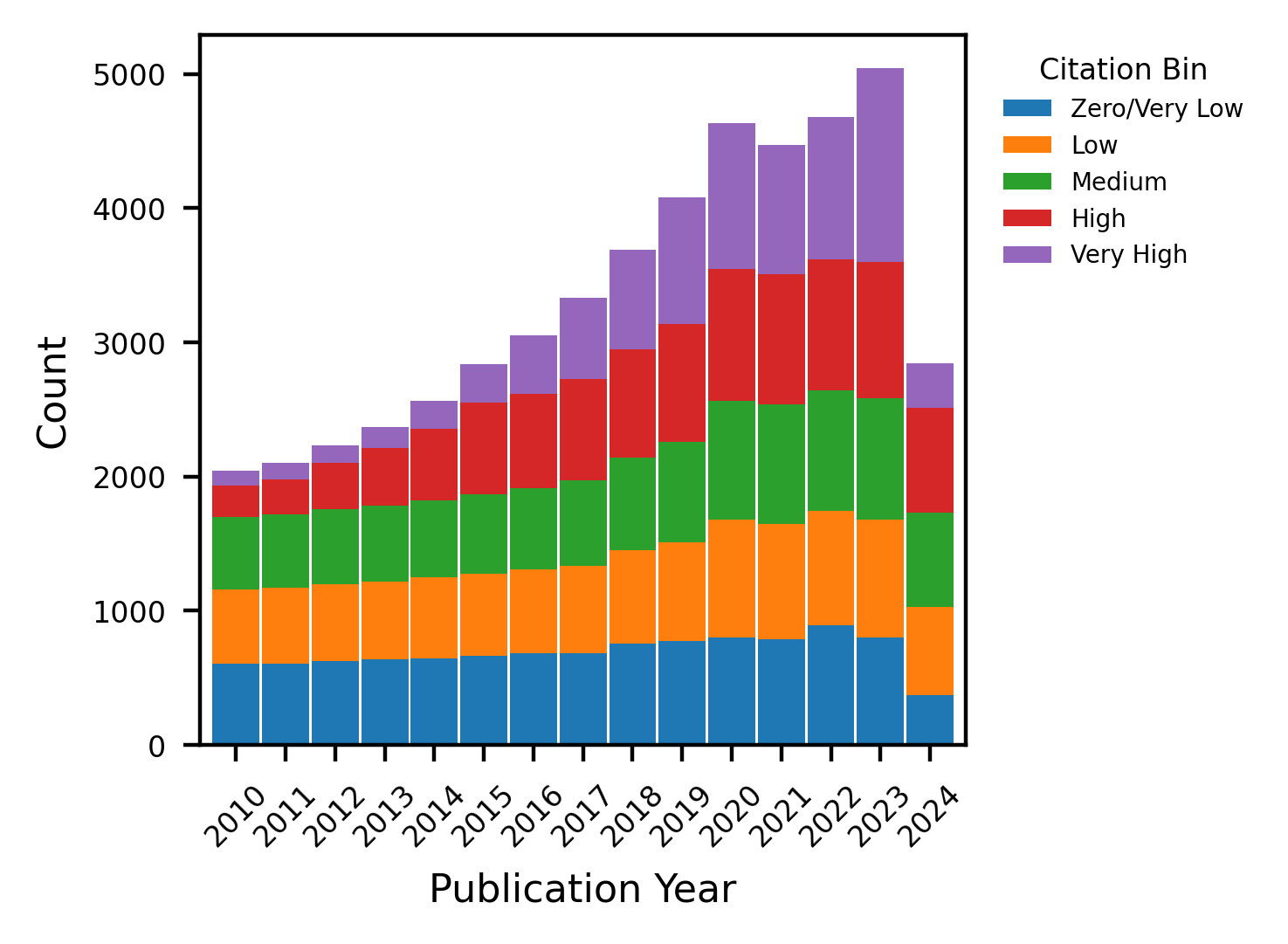}
    \caption{
    Publication year distribution of the reward dataset across citation-impact bins.
    The dataset spans a broad temporal range and contains papers from diverse citation-impact categories, demonstrating balanced impact coverage across publication years.
    }
    \label{fig:reward_year_bin_distribution}
\end{figure}

\begin{figure}[t]
    \centering
    \includegraphics[width=\linewidth]{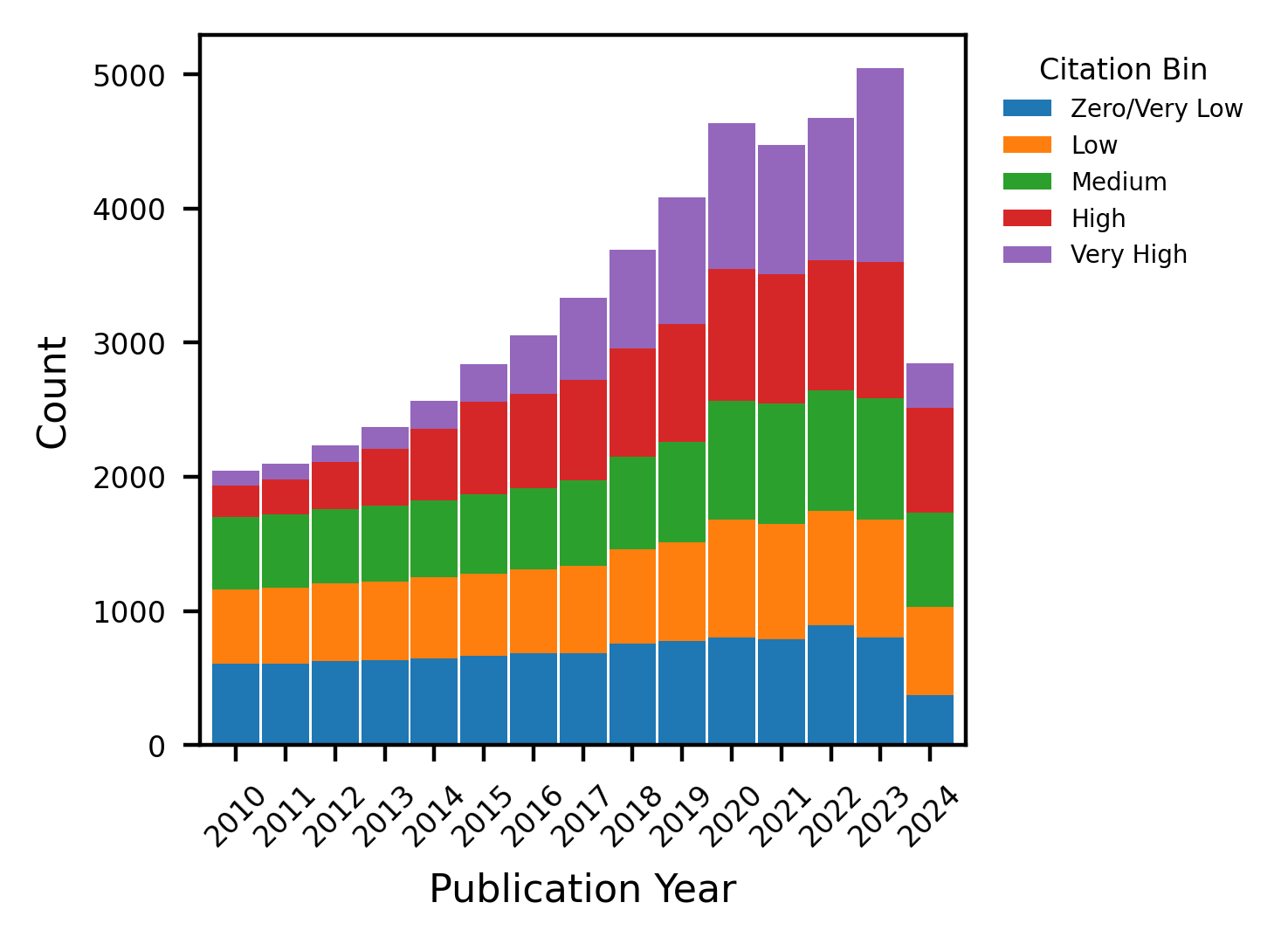}
    \caption{
    Publication year distribution of the SFT dataset across citation-impact bins.
    The dataset exhibits broad temporal coverage with citation-impact diversity distributed consistently across years.
    }
    \label{fig:sft_year_bin_distribution}
\end{figure}

\begin{table}[t]
\centering
\scriptsize
\renewcommand{\arraystretch}{1.15}
\setlength{\tabcolsep}{4pt}

\begin{tabular}{p{0.52\linewidth}p{0.33\linewidth}}
\toprule
\textbf{Training / Evaluation Parameter} & \textbf{Value} \\
\midrule

Base Model & Qwen3-8B \\
RL Algorithm & DAPO \\
Reward Model & Qwen3 8B with CORN \\
Maximum Sequence Length & 2048 \\
Maximum Completion Length & 256 \\
Evaluation Generation Length & 500 \\
No of epoch & 1 \\
Per-device Batch Size & 1 \\
Gradient Accumulation Steps & 4 \\
Effective Batch Size & 4 \\
Number of Generations & 4 \\
Learning Rate & $1 \times 10^{-6}$ \\
LoRA Rank ($r$) & 64 \\
LoRA Alpha & 128 \\
LoRA Dropout & 0.1 \\
Sampling Strategy & Stochastic \\
Temperature & 0.9 \\
Top-p & 0.95 \\
Repetition Penalty & 1.05 \\
Reward Scaling Range & $[0,5]$ \\
Reward Scaling Method & Group Norm. \\
Random Seed & 42 \\
GPU & NVIDIA A100 80GB \\
\bottomrule
\end{tabular}
\vspace{0.3cm}
\caption{
Training and evaluation configuration used for GRPO-based impact-aligned research idea generation.
}
\label{tab:training_eval_parameters}
\end{table}

\begin{table}[t]
\centering
\small
\renewcommand{\arraystretch}{1.2}
\setlength{\tabcolsep}{6pt}

\begin{tabular}{lc}
\toprule
\textbf{Evaluation Parameter} & \textbf{Value} \\
\midrule

Base Model & Qwen3-8B \\
Maximum Sequence Length & 2048 \\
Maximum New Tokens & 500 \\
Batch Size & 128 \\
Sampling Strategy & Stochastic Sampling \\
Temperature & 0.9 \\
Top-p & 0.95 \\
Repetition Penalty & 1.05 \\
Random Seed & 42 \\
Evaluation Split & SFT/RL Test Set \\
Generation Format & Single Paragraph \\
Output Length Constraint & 8--12 Sentences \\
GPU & NVIDIA A100 80GB PCIe \\
\bottomrule
\end{tabular}
\vspace{0.3cm}
\caption{
Generation and evaluation configuration used for idea generation experiments.
All models were evaluated using identical decoding and inference settings to ensure fair comparison across baseline, SFT, and RL models.
Experiments were conducted on a single NVIDIA A100 80GB PCIe GPU.
}
\label{tab:eval_parameters}
\end{table}

\onecolumn

\subsection{Ethical Considerations}

 There is a potential misuse risk in deploying impact-oriented ideation systems for gaming academic incentives. For example, such models could be used to produce ideas optimized for attention, publication likelihood, or anticipated citation uptake rather than epistemic value, reproducibility, or long-term societal benefit. We therefore do not view citation-aligned generation as a substitute for scientific judgment, but rather as one tool among many for exploring possible directions.

\subsection{Prompts}

\subsubsection{Prompt A}

\begin{tcblisting}{
    enhanced,
    breakable,
    colback=gray!5,
    colframe=black,
    title={Evaluation Prompt used to select best prompt for Research Goal and Idea Extraction},
    fonttitle=\bfseries,
    listing only,
    label={lst:Prompt_evalualtion}
}
You are an expert evaluator of scientific information extraction quality.

Your task is to evaluate seven anonymized candidate outputs generated from the same research paper.
Each candidate output contains:
- research_goal
- idea_description

You must judge each candidate ONLY against the provided paper title and full paper text.

==================================================
EVALUATION TARGET
==================================================

Each candidate output should extract two things from the paper:

1. research_goal
This should describe:
- the scientific problem addressed by the paper
- the research objective
- the task framing
- the input/output behavior if stated
- the motivation only if explicitly stated in the paper

It should focus on the objective/problem, not the internal technical method.

2. idea_description
This should describe:
- the core technical proposal
- how it works mechanistically
- representation, modules, architecture, or stages
- equations, losses, objectives, or assumptions if present
- training/inference procedure if present

It should focus on the technical mechanism, not results, comparisons, or benefits.

==================================================
EXCLUDED CONTENT
==================================================

Candidate outputs should NOT include:
- empirical results
- performance metrics as outcomes
- comparisons to baselines or prior work
- claims of improvement or superiority
- advantages or benefits
- limitations or failure cases
- future work
- related work summaries

==================================================
SCORING DIMENSIONS
==================================================

For each candidate output, assign a score from 1 to 5 for each dimension:

1. research_goal_fidelity
2. idea_fidelity
3. mechanistic_completeness
4. goal_method_separation
5. exclusion_compliance
6. technical_specificity
7. coherence
8. overall_quality

Scoring scale:
5 = Excellent
4 = Good
3 = Acceptable / Mixed
2 = Weak
1 = Poor / Incorrect

==================================================
JUDGING RULES
==================================================

- Be strict about hallucinated technical details.
- Do not reward fluent but unsupported content.
- Penalize outputs that mix research objective with technical mechanism.
- Penalize outputs that include excluded content.
- Prefer outputs that are faithful, mechanistic, and specific to the paper.
- Do not use outside knowledge; rely only on the paper text and candidate outputs.

==================================================
INPUT
==================================================

Paper Title:
{title}

Full Paper Text:
{full_text}

Candidate Outputs:

Output A
research_goal:
{research_goal_A}

idea_description:
{idea_description_A}

Output B
research_goal:
{research_goal_B}

idea_description:
{idea_description_B}

Output C
research_goal:
{research_goal_C}

idea_description:
{idea_description_C}

Output D
research_goal:
{research_goal_D}

idea_description:
{idea_description_D}

Output E
research_goal:
{research_goal_E}

idea_description:
{idea_description_E}

Output F
research_goal:
{research_goal_F}

idea_description:
{idea_description_F}

Output G
research_goal:
{research_goal_G}

idea_description:
{idea_description_G}

==================================================
OUTPUT FORMAT
==================================================

Return ONLY valid JSON in the following format:

{{
  "title": "{title}",
  "scores": {{
    "A": {{
      "research_goal_fidelity": 1,
      "idea_fidelity": 1,
      "mechanistic_completeness": 1,
      "goal_method_separation": 1,
      "exclusion_compliance": 1,
      "technical_specificity": 1,
      "coherence": 1,
      "overall_quality": 1
    }},
    "B": {{
      "research_goal_fidelity": 1,
      "idea_fidelity": 1,
      "mechanistic_completeness": 1,
      "goal_method_separation": 1,
      "exclusion_compliance": 1,
      "technical_specificity": 1,
      "coherence": 1,
      "overall_quality": 1
    }},
    "C": {{
      "research_goal_fidelity": 1,
      "idea_fidelity": 1,
      "mechanistic_completeness": 1,
      "goal_method_separation": 1,
      "exclusion_compliance": 1,
      "technical_specificity": 1,
      "coherence": 1,
      "overall_quality": 1
    }},
    "D": {{
      "research_goal_fidelity": 1,
      "idea_fidelity": 1,
      "mechanistic_completeness": 1,
      "goal_method_separation": 1,
      "exclusion_compliance": 1,
      "technical_specificity": 1,
      "coherence": 1,
      "overall_quality": 1
    }},
    "E": {{
      "research_goal_fidelity": 1,
      "idea_fidelity": 1,
      "mechanistic_completeness": 1,
      "goal_method_separation": 1,
      "exclusion_compliance": 1,
      "technical_specificity": 1,
      "coherence": 1,
      "overall_quality": 1
    }},
    "F": {{
      "research_goal_fidelity": 1,
      "idea_fidelity": 1,
      "mechanistic_completeness": 1,
      "goal_method_separation": 1,
      "exclusion_compliance": 1,
      "technical_specificity": 1,
      "coherence": 1,
      "overall_quality": 1
    }},
    "G": {{
      "research_goal_fidelity": 1,
      "idea_fidelity": 1,
      "mechanistic_completeness": 1,
      "goal_method_separation": 1,
      "exclusion_compliance": 1,
      "technical_specificity": 1,
      "coherence": 1,
      "overall_quality": 1
    }}
  }},
  "ranking_best_to_worst": ["A", "B", "C", "D", "E", "F", "G"],
  "best_output": "A",
  "worst_output": "G",
  "summary": "Short paragraph explaining why the best output is strongest and why the worst output is weakest."
}}

\end{tcblisting}

\subsubsection{Prompt B}

\begin{tcblisting}{
    enhanced,
    breakable,
    colback=gray!5,
    colframe=black,
    title={Research Goal and Idea Extraction Prompt},
    fonttitle=\bfseries,
    listing only,
    label={lst:research_goal_idea_prompt}
}


You are a senior research scientist tasked with converting full-paper research content into two structured outputs:
1. the paper's research_goal
2. the paper's idea_description

Your extraction must rely primarily on the full paper text, using the title and abstract only as supporting context. Focus exclusively on the scientific objective, technical problem, task formulation, proposed method, contribution, and mechanism of what is introduced in the paper. Ignore empirical results, performance comparisons, claimed advantages, limitations, failure cases, related work, background discussion, and future work.

INPUT
Corpus ID: {corpus_id}
Title: {title}
Abstract: {abstract}
Full Paper Text: {full_text}

INTERNAL RECONSTRUCTION INSTRUCTIONS
Before generating the final answer, internally reconstruct the paper's scientific content by identifying and organizing, when explicitly present in the full paper:
- the core research objective,
- the central scientific or technical problem,
- the task setting,
- the input data type, modality, representation, or observation space,
- the desired output, prediction target, generated artifact, or capability,
- the scientific motivation if explicitly stated,
- the core technical contribution,
- the detailed mechanism of the proposal from input to output,
- the theoretical assumptions, principles, definitions, or formal foundations,
- the architectural structure, modules, subcomponents, layers, and transformations,
- the interaction among components and the information flow through the system,
- the mathematical formulations, equations, variables, symbols, constraints, and formal relationships,
- the objective functions, losses, optimization targets, regularization terms, or probabilistic structure,
- the algorithmic procedures for training, optimization, inference, prediction, or decoding,
- the implementation details necessary for understanding how the method operates.

Do not reveal this intermediate reconstruction. Use it only to produce the final output.

SECTION PRIORITY
When reconstructing the method, prioritize content from sections that define the proposed approach, model, architecture, formulation, algorithm, objective, training, inference, implementation, or method details. Use the introduction and abstract mainly to clarify the problem setting and motivation, but do not let them replace methodological detail found later in the full paper.

FIELD DEFINITIONS

research_goal:
Write one substantial paragraph of 4 to 6 sentences that explains the core research objective of the work. It must state the scientific or technical problem being addressed, the task setting, the type of input data or observations involved, the expected output or capability, and the motivation only if it is explicitly stated in the full paper. If the paper explicitly frames the task through a benchmark, dataset, or evaluation setup and that framing is essential to defining the objective, incorporate it naturally. This paragraph must remain focused on the objective and problem formulation rather than the method. Do not use generic paper-centric phrasing such as "This paper aims", "The paper addresses", or "The paper proposes". Use research-centric phrasing such as "The core objective of this research is", "The primary objective is", or "The research investigates how to".

idea_description:
Write one detailed scientific paragraph of 8 to 12 sentences that explains the proposed idea, technical contribution, methodology, and mechanism as a cohesive process from input to output. This paragraph must not be a short summary. It must reconstruct how the method works step by step, including representations, preprocessing assumptions if stated, modules, subcomponents, architectural structure, transformations, message passing or computation flow, intermediate states, formal definitions, equations, variables, losses, constraints, probabilistic assumptions, optimization objectives, training procedure, inference procedure, and implementation details whenever these are explicitly present in the full paper. When multiple components are present, explain how they interact, what each component consumes and produces, and how the overall system produces the final output. If the paper defines objective functions, equations, symbolic variables, or optimization targets, explain them in words and include the role of the variables when stated. If some details are only partially specified, provide the most complete defensible reconstruction possible without inventing missing information.

DEPTH AND SPECIFICITY REQUIREMENTS
- research_goal must be specific, content-rich, and grounded in the actual task described in the paper.
- idea_description must be mechanistic and technically detailed, not a high-level overview.
- Avoid generic statements such as "a novel framework is introduced", "the method improves performance", "the approach is effective", or "the model captures rich features" unless immediately followed by concrete scientific detail grounded in the full text.
- Do not compress the method into vague abstractions if the full paper provides operational detail.
- Prefer explicit scientific language over generalized summarization.
- Explain the flow of information through the proposed system whenever the paper provides enough detail.
- If training and inference are distinct, describe both separately.
- If the method includes multiple stages, explain them in order.
- If mathematical formulations are present, include them conceptually and explain their function rather than merely mentioning that equations exist.

EXTRACTION RULES
- Use only content grounded in the title and full paper text, with full paper text taking precedence.
- Focus only on the paper's own proposal and its mechanism.
- Reconstruct the method as a unified scientific process rather than a list of disconnected facts.
- Include technical details only when they are supported by the full paper.
- If the proposal name is noisy, abbreviated, inconsistent, implementation-qualified, or not scientifically informative, infer the underlying technical content from the full text rather than relying on the name.
- Do not introduce claims or interpretations that are not directly supported by the paper.

EXCLUSIONS
Exclude all of the following:
- empirical findings,
- performance metrics,
- benchmark scores,
- comparison language,
- claims of superiority over prior work,
- claimed benefits or advantages,
- limitations, drawbacks, or failure cases,
- future work suggestions,
- related work discussion,
- historical background,
- qualitative evaluation discussion,
- experimental setup unless essential for defining the research goal,
- names given to the model, method, or framework inside the explanatory prose unless necessary for identification in the output,
- unsupported speculation.

WRITING RULES
- Return exactly one valid JSON object and nothing else.
- Do not use bullet points.
- Do not use markdown.
- Do not add commentary outside the JSON.
- Use natural, precise, cohesive scientific prose.
- research_goal must be a single paragraph of 4 to 6 sentences.
- idea_description must be a single paragraph of 8 to 12 sentences and should generally be substantially longer than research_goal.
- Both fields must be concrete and content-specific; generic boilerplate phrasing is not acceptable.
- If the paper contains rich technical detail, the output must preserve that detail in compressed but explicit scientific language.

OUTPUT
Return exactly one valid JSON object only:

{{
  "paper_id": {corpus_id},
  "title": "{title}",
  "research_goal": "single specific paragraph grounded only in the paper",
  "idea_description": "single detailed scientific paragraph grounded only in the paper"
}}

\end{tcblisting}

\subsubsection{Prompt C}
\begin{tcblisting}{
    enhanced,
    breakable,
    colback=gray!5,
    colframe=black,
    title={Generation Prompt Used for Baseline, SFT, RL},
    fonttitle=\bfseries,
    listing only,
    label={lst:generation_prompt}
}

System Prompt:
You are an expert AI research assistant. Given a research goal, write the idea_description as technical scientific prose. Do not include evaluation, comparisons to prior work, benchmarks, or performance/advantage claims. Avoid repetition; do not restate the same sentence with minor rewording.

User Prompt:
Research Goal:
<research_goal>

Write the idea_description as ONE paragraph of 8-12 sentences. It must explain the method mechanistically from input to output, including (when applicable): representations, components/modules and information flow, objective/loss, training procedure, and inference procedure.

Constraints:
- No bullet points, no headings, no markdown.
- Do not use claim language like: novel, significant, advancement, outperforms, state-of-the-art, better, improves, superior.
- Do not include expected outcomes or evaluation discussion.
- Do not repeat sentences or phrases; each sentence must add new technical information.

Return only the idea_description paragraph.

\end{tcblisting}

\subsubsection{Prompt D}
\begin{tcblisting}{
    enhanced,
    breakable,
    colback=gray!5,
    colframe=black,
    title={Evaluator Prompt Used for Baseline, SFT, and RL Model Assessment},
    fonttitle=\bfseries,
    listing only,
    label={lst:evaluator_prompt}
}


System Prompt:
You are an expert scientific evaluator.

Your task is to evaluate whether a generated research idea is scientifically impactful relative to a ground-truth reference idea for the same research goal.

This is a conservative binary evaluation.

The goal is NOT to judge textual similarity. A generated idea may differ from the reference idea. However, it should receive label = 1 only if it is clearly relevant, technically plausible, sufficiently specific, non-obvious, and expected to make a scientific contribution that is comparable to or greater than the ground-truth reference idea.

Use the ground-truth ordinal impact label, if provided, as contextual information about the reference idea's expected impact. If the reference idea has a high ordinal impact label, the generated idea must meet a similarly high standard to receive label = 1.

Evaluation criteria:
1. Relevance to the research goal
2. Specificity and clarity
3. Novelty beyond standard or obvious approaches
4. Technical soundness
5. Feasibility
6. Scientific contribution
7. Relative impact compared with the ground-truth reference idea

Decision policy:
- Assign label = 1 only if the generated idea is clearly impactful relative to the reference idea.
- Assign label = 0 if the generated idea is irrelevant, vague, generic, obvious, technically weak, infeasible, redundant without meaningful added value, or substantially weaker than the reference idea.
- Assign label = 0 if the generated idea is merely plausible but lacks a clear mechanism, method, hypothesis, or scientific contribution.
- Assign label = 0 if the generated idea only restates the research goal or proposes routine extensions without meaningful novelty.
- Assign label = 0 if the generated idea is complementary but materially less impactful than the reference idea.
- If uncertain or borderline, assign label = 0.

Important calibration:
- Do not reward broad ambition alone.
- Do not infer missing technical details.
- Do not assume novelty unless it is evident from the idea.
- Do not mark an idea impactful just because it is relevant.
- The generated idea must be comparable to the reference idea in expected scientific value, not merely related to the same topic.

Return only valid JSON. Do not include markdown. Do not include any additional text outside the JSON.

The JSON output must follow exactly this schema:
{
  "decision": "Impactful" or "Not Impactful",
  "label": 1 or 0,
  "rationale": "A concise explanation of why the generated idea is or is not impactful relative to the ground-truth idea.",
  "criterion_assessment": {
    "relevance_to_goal": "brief assessment",
    "specificity_and_clarity": "brief assessment",
    "novelty": "brief assessment",
    "technical_soundness": "brief assessment",
    "feasibility": "brief assessment",
    "scientific_contribution": "brief assessment",
    "relative_impact_against_reference": "brief assessment"
  }
}

User Prompt:
Evaluate the following generated research idea.

Research Goal:
<research_goal>

Ground-Truth Reference Idea Impact Label:
- Ordinal class: <ground_truth_ordinal_class>
- Ordinal index: <ground_truth_ordinal_index>

Ground-Truth Reference Idea:
<ground_truth_idea>

Generated Idea:
<generated_idea>

Your task:
Determine whether the Generated Idea is scientifically impactful relative to the Ground-Truth Reference Idea under the same Research Goal.

Important:
- Do not judge by textual similarity alone.
- Do not mark the Generated Idea as impactful merely because it is relevant.
- Label = 1 only if the Generated Idea is specific, technically plausible, non-obvious, and comparable to or stronger than the Ground-Truth Reference Idea in expected scientific contribution.
- Label = 0 if the Generated Idea is vague, generic, obvious, weakly justified, routine, redundant, or clearly less impactful than the Ground-Truth Reference Idea.
- If the decision is uncertain or borderline, choose label = 0.

Return only valid JSON with:
- decision
- label
- rationale
- criterion_assessment

\end{tcblisting}

\begin{table*}[!t]
\centering

\begin{tcolorbox}[
enhanced,
width=\textwidth,
colback=white,
colframe=black,
fonttitle=\bfseries,
sharp corners,
boxrule=0.6pt,
left=1.5pt,
right=1.5pt,
top=1.5pt,
bottom=1.5pt
]

\footnotesize
\setlength{\tabcolsep}{2.0pt}
\renewcommand{\arraystretch}{0.98}

\textbf{Research Goal:}

\vspace{1pt}
The core objective of this research is to enhance few-shot image classification by training a more generalized embedding network using self-supervised learning (SSL). Few-shot classification is motivated by the challenge of accurately classifying unseen classes with a limited number of labeled samples. The focus is on developing an embedding network that can effectively generalize to new tasks based on minimal data, overcoming the limitations imposed by traditional supervised learning methods. This work aims to leverage the abundance of unlabeled data to improve the quality of embeddings used in few-shot classification tasks.

\vspace{0.18cm}
\textbf{Color Legend:}

\vspace{1pt}
\baselinehl{Baseline} weak/abstract technical grounding \hspace{0.25cm}
\sfthl{SFT} moderate structured grounding \hspace{0.25cm}
\rlhl{RL} implementation-oriented grounding

\vspace{0.18cm}

\begin{tabularx}{\textwidth}{
>{\raggedright\arraybackslash}p{1.0cm}
>{\raggedright\arraybackslash}X
}
\toprule
\textbf{Method} & \textbf{Generated Idea Description} \\
\midrule

\textbf{Baseline} &
This research proposes a self-supervised learning approach to train an embedding network for few-shot image classification by leveraging unlabeled data to generate more generalized representations. The process begins with input images being passed through a series of convolutional layers that extract hierarchical visual features. These features are then transformed into a lower-dimensional embedding space via a projection head, which includes several fully connected layers with non-linear activation functions. \baselinehl{To encourage robust feature learning, the model is trained using a contrastive loss function that maximizes the similarity between embeddings of images from the same class while minimizing the similarity between those from different classes.} During training, images are augmented through random transformations such as rotation, flipping, and color jittering to increase the diversity of the input distribution. The augmented views are treated as positive pairs, and the model learns to align their embeddings in the latent space. \baselinehl{Additionally, a queue-based memory mechanism stores negative examples to provide long-term contrastive supervision.} The learned embeddings are then used in a few-shot classification setup where support sets are used to compute prototype vectors for each class. During inference, a query image is embedded and compared to these prototypes using a similarity metric such as cosine distance. \baselinehl{The class with the highest similarity score is predicted as the output label.} This method enables the model to generalize effectively to unseen classes with minimal labeled data by capitalizing on the rich structure of unlabeled data. The overall framework integrates representation learning and task-specific adaptation to improve generalization in low-data scenarios. \\
\midrule

\textbf{SFT} &
The proposed method, SSL4Few, utilizes a two-step training approach to construct a generalizable embedding network for few-shot classification. Initially, it trains a backbone model using SSL on a large-scale dataset of available image pairs through contrastive loss, aiming to learn robust feature representations while maintaining a balance between similarity for the same images and dissimilarity for different classes. \sfthl{Subsequently, during fine-tuning on specific few-shot tasks, SSL4Few introduces a unique loss function combining both SSL loss and cross-entropy loss, allowing the embedding network to adaptively adjust its weight towards either SSL or supervised learning based on task difficulty.} When the model identifies itself as facing an easy task, it emphasizes SSL, thereby retaining the generalization capability acquired during pre-training. \sfthl{Conversely, for harder tasks where SSL alone may not provide sufficient discriminative power, the model switches to minimizing cross-entropy loss to maximize classification accuracy.} This dynamic adjustment of losses ensures optimal performance across varying difficulty levels, culminating in a robust embedding space that enhances few-shot classification capabilities without relying on extensive labeled data. \sfthl{The methodology thus bridges the gap between SSL and supervised learning, demonstrating how they can complement each other in achieving better performance on few-shot learning tasks.} \\
\midrule

\textbf{RL} &
This paper introduces a meta-learning framework for SSL that trains a pre-trained embedding network to maximize its generalization across diverse downstream few-shot tasks. \rlhl{It employs three innovative components: a hard negative mining strategy to reinforce the contrast between positive support instances and their hardest negatives; task-specific augmentation policies that adaptively adjust data augmentations for different tasks, promoting diverse feature representations; and a cross-task contrastive loss to align features across multiple tasks during training.} The process begins with SSL pre-training where the embedding network learns to distinguish between labeled positives and negatives through hard negative mining, ensuring that each task's positive set is well separated from a broader set of negative instances drawn from unlabeled data. \rlhl{Subsequently, task-specific augmentations are optimized during fine-tuning based on the specific characteristics of each downstream few-shot task.} The overall architecture minimizes a loss function combining both SSL and cross-task contrastive losses, enabling the system to learn robust, semantically meaningful features that facilitate high accuracy in few-shot classification. \rlhl{By jointly optimizing these strategies throughout the entire training phase, the approach aims to improve the network’s ability to extract useful information from minimal labeled examples while leveraging extensive unlabeled data.} \\
\bottomrule
\end{tabularx}

\vspace{0.18cm}

\noindent\textbf{Observation:}
The RL-generated idea is identified as impactful because it proposes a concrete, non-trivial meta-learning framework that explicitly tailors self-supervised embeddings for few-shot tasks through three specific mechanisms: hard negative mining, task-specific augmentation policies, and a cross-task contrastive loss. Compared to the Baseline idea, which follows a standard contrastive SSL and prototypical few-shot classification pipeline, the RL approach goes beyond routine reuse of known components by introducing task-aware mechanisms for improving generalization. Compared to the SFT idea, which mainly combines SSL pretraining with a heuristic task-dependent weighting between contrastive and cross-entropy losses, the RL approach provides more clearly specified and methodologically richer components. Its use of hard negative mining strengthens class separation, task-specific augmentation policies adapt representation learning to downstream few-shot tasks, and cross-task contrastive learning aligns features across multiple tasks. These elements make the RL idea more specific, innovative, and likely to yield meaningful scientific contribution, whereas the Baseline and SFT ideas are judged less impactful because they are more incremental and under-specified.

\end{tcolorbox}

\vspace{2pt}
\caption{Qualitative Comparison of Baseline, SFT, and RL Generated Research Ideas}
\label{tab:qualitative_comparison_ssl_fewshot}

\end{table*}

\begin{table*}[!t]
\centering

\begin{tcolorbox}[
enhanced,
width=\textwidth,
colback=white,
colframe=black,
fonttitle=\bfseries,
sharp corners,
boxrule=0.6pt,
left=1.8pt,
right=1.8pt,
top=1.8pt,
bottom=1.8pt
]

\fontsize{9pt}{9.2pt}\selectfont
\setlength{\tabcolsep}{2.8pt}
\renewcommand{\arraystretch}{0.94}

\textbf{Research Goal:}

\vspace{1pt}
The core objective of this research is to enhance emergency response systems by leveraging Large Language Models (LLMs) to identify and classify emergency situations through social media posts and direct emergency messages. The task involves processing unstructured data from users describing their emergencies in real-time. The expected output includes actionable insights that improve situational awareness for first responders and provide instructions to affected individuals during crisis situations. Through this approach, the work seeks to address challenges associated with traditional emergency management systems, such as information overload and communication bottlenecks.

\vspace{0.15cm}
\textbf{Color Legend:}

\vspace{1pt}
\baselinehl{Baseline} weak/abstract technical grounding \hspace{0.22cm}
\sfthl{SFT} moderate structured grounding \hspace{0.22cm}
\rlhl{RL} implementation-oriented grounding

\vspace{0.15cm}

\begin{tabularx}{\textwidth}{
>{\raggedright\arraybackslash}p{1.5cm}
>{\raggedright\arraybackslash}X
}
\toprule
\textbf{Method} & \textbf{Generated Idea Description} \\
\midrule

\textbf{Baseline} &
This research proposes a system that utilizes Large Language Models to analyze and categorize emergency-related content from social media platforms and direct messaging channels. The process begins by ingesting raw textual data, which is preprocessed to remove noise and standardize formats, ensuring consistency across diverse input sources. Preprocessed text is then encoded into contextualized embeddings using a transformer-based architecture, capturing semantic relationships within the input. \baselinehl{These embeddings are fed into a classification module, which employs a multi-layered neural network to distinguish between different types of emergency scenarios based on learned patterns from labeled training data.} The model incorporates attention mechanisms to prioritize critical linguistic features that are indicative of emergency conditions. During training, the model is optimized using a cross-entropy loss function that minimizes prediction errors across all emergency categories. The training dataset consists of annotated examples collected from historical emergency reports and manually curated social media posts. \baselinehl{Inference proceeds by transforming incoming text into embeddings and passing them through the trained network to generate probability distributions over possible emergency classifications.} The output includes structured metadata that highlights the identified emergency type, severity level, and potential location, enabling real-time decision-making. \baselinehl{The system also integrates temporal filtering to ensure relevance and timeliness of the classified information.} All components operate in an end-to-end pipeline designed for low-latency processing in dynamic crisis environments. This approach enables automated extraction of actionable intelligence from unstructured user-generated content, supporting rapid response coordination. \\
\midrule

\textbf{SFT} &
The proposed system, ER-Mind, operates by integrating LLMs at multiple stages of emergency response management. Initially, when an emergency occurs, the system utilizes pre-trained LLMs, such as GPT-3.5, which are fine-tuned on a dataset comprising both social media posts and emergency messages. \sfthl{This allows the models to effectively identify and classify emergencies based on the contextual clues provided by users.} Once an emergency is recognized, LLMs generate relevant instructions tailored for the situation, ensuring the information delivered is accurate and applicable. \sfthl{Additionally, during post-emergency scenarios, the LLMs facilitate community engagement by summarizing feedback from survivors or victims, promoting collective recovery efforts.} The architecture supports real-time interactions with first responders by enabling quick access to essential information, thereby streamlining the response process. \sfthl{Through continuous interaction with data from various sources, ER-Mind aims to adaptively improve its capabilities over time, ultimately supporting efficient emergency response strategies.} \\
\midrule

\textbf{RL} &
The proposed approach introduces a framework called EMERGENCY, which utilizes pre-trained large language models for the automatic extraction of emergency events and contextual information from both text-based messages and images shared during crises. \rlhl{Initially, the system processes incoming emergency messages by using a prompt engineering mechanism designed to elicit high-quality outputs from LLMs, ensuring accurate detection of various types of emergencies, including natural disasters and accidents.} A dual encoder network is then employed to embed both the extracted event descriptions and the corresponding multimodal features into a unified representation space, enabling efficient semantic retrieval and clustering of similar emergencies. \rlhl{To mitigate the hallucination risks inherent in LLMs, the architecture incorporates a fine-tuning module that uses a hybrid loss function, balancing between maximizing the cosine similarity of the embeddings while penalizing hallucinated data through cross-entropy loss.} This results in a robust embedding that is capable of representing different facets of emergencies, such as their severity and context, facilitating the aggregation of emergency signals across diverse sources. \rlhl{Furthermore, the system includes a knowledge retrieval component that enhances the accuracy of emergency classification and supports real-time updates with relevant public safety resources, thereby improving the overall efficacy of emergency response efforts.} \\
\bottomrule
\end{tabularx}

\vspace{0.15cm}

\noindent\textbf{Observation:}
The RL-generated idea is identified as impactful because it proposes a concrete, technically detailed framework that goes beyond generic LLM-based emergency classification. Compared to the Baseline idea, which mainly describes a standard supervised text-classification pipeline using transformer embeddings, neural classification, cross-entropy training, and temporal filtering, the RL approach introduces a richer multimodal and retrieval-oriented architecture. Compared to the SFT idea, which fine-tunes LLMs for emergency classification, instruction generation, and feedback summarization but remains high-level, the RL approach provides more specific methodological contributions. It integrates prompt-engineered LLMs with a dual-encoder multimodal embedding space, semantic retrieval and clustering, hallucination-aware hybrid loss, and knowledge retrieval for real-time public safety guidance. These components directly address important emergency-response challenges such as multimodal crisis understanding, hallucination mitigation, situational aggregation, severity/context representation, and real-time guidance. Therefore, the RL idea is more innovative, technically grounded, and likely to yield stronger scientific contribution than the Baseline and SFT ideas, which are judged less impactful because they rely more on routine or under-specified LLM application pipelines.

\end{tcolorbox}

\vspace{2pt}
\caption{Qualitative Comparison of Baseline, SFT, and RL Generated Research Ideas}
\label{tab:qualitative_comparison_emergency_llm}

\end{table*}

\begin{table*}[!t]
\centering

\begin{tcolorbox}[
enhanced,
width=\textwidth,
colback=white,
colframe=black,
fonttitle=\bfseries,
sharp corners,
boxrule=0.8pt
]

\textbf{Research Goal:}

\vspace{2pt}
The core objective of this research is to develop an automatic method for pavement crack detection to enhance road safety. The work specifically addresses the challenges posed by the intensity inhomogeneity of cracks and the complexities in the background of pavement images, where low contrast and shadows may obscure cracks. The task involves generating a robust prediction of crack locations from input images using advanced deep learning techniques. This work aims to alleviate the labor-intensive manual detection process by implementing a more efficient automated solution.

\vspace{0.3cm}
\textbf{Color Legend:}

\vspace{2pt}
\baselinehl{Baseline} weak/abstract technical grounding \hspace{0.35cm}
\sfthl{SFT} moderate structured grounding \hspace{0.35cm}
\rlhl{RL} implementation-oriented grounding

\vspace{0.3cm}

\small
\setlength{\tabcolsep}{4pt}
\renewcommand{\arraystretch}{1.05}

\begin{tabularx}{\textwidth}{
>{\raggedright\arraybackslash}p{1.7cm}
>{\raggedright\arraybackslash}X
}
\toprule
\textbf{Method} & \textbf{Generated Idea Description} \\
\midrule

\textbf{Baseline} &
The proposed method employs a convolutional neural network designed to detect pavement cracks by addressing the challenges of intensity inhomogeneity and complex backgrounds. The model extracts hierarchical features from pavement images and integrates an attention module to emphasize crack regions while suppressing irrelevant background information. \baselinehl{A multi-scale feature fusion strategy combines outputs from different network depths to improve robustness across varying crack widths and orientations.} \baselinehl{The loss function incorporates a weighted combination of binary cross-entropy and dice coefficient to handle class imbalance and improve segmentation accuracy.} The architecture follows an encoder-decoder design that preserves spatial resolution during feature propagation and generates a probability map highlighting potential crack locations. \baselinehl{Post-processing with non-maximum suppression is applied to refine predictions and remove overlapping detections.} \\
\midrule

\textbf{SFT} &
The proposed method leverages a fully convolutional network (FCN) architecture to predict pavement crack locations directly from input images. Preprocessing techniques such as histogram equalization and Sobel operators are used to enhance contrast and identify edge information associated with cracks. \sfthl{An attention mechanism is incorporated into the FCN model to amplify relevant crack features while suppressing irrelevant regions.} The architecture combines convolutional layers with a feature fusion module to capture cracks at multiple scales. \sfthl{The output layer produces a binary probability map indicating crack likelihood at each pixel location.} During training, cross-entropy loss and data augmentation methods such as flipping and scaling are employed to improve generalization across varying pavement conditions. \sfthl{Inference is performed end-to-end, producing crack localization outputs suitable for automated maintenance analysis.} \\
\midrule

\textbf{RL} &
The proposed methodology introduces an attention-based image pyramid to address low-resolution inputs and scale variations in pavement crack detection. \rlhl{The pyramid consists of a shallow branch for high-resolution feature extraction and a deeper hourglass branch for capturing fine crack details.} To reduce interference from non-crack regions during upsampling, \rlhl{a crack-specific adaptive filter is integrated within the attention branch to refine feature representations.} The architecture further incorporates a dual-branch dilated convolutional network (Dilated-DBCRNet), where a global feature extraction module processes both original and warped images for contextual understanding, while a local feature extraction module utilizes residual connections, squeeze-and-excitation blocks, and dilated convolutions to capture edge-level crack structures. \rlhl{The network aggregates multi-scale contextual and local features to improve crack localization under complex background conditions.} A joint binary cross-entropy and Dice loss function is used for end-to-end optimization, producing accurate pixel-wise crack predictions across diverse pavement scenarios. \\
\bottomrule
\end{tabularx}

\normalsize

\vspace{0.3cm}

\noindent\textbf{Observation:}
The RL-generated idea is identified as impactful because it introduces a more specialized and technically integrated framework compared to the baseline and SFT approaches. Unlike the baseline encoder-decoder design and the SFT FCN-based formulation, the RL approach combines an attention-based image pyramid, crack-specific adaptive filtering, and a dual-branch dilated convolutional architecture for improved multi-scale feature modeling and contextual reasoning. The integration of hourglass structures, warped-image contextual processing, residual connections, squeeze-and-excitation blocks, and dilated convolutions provides stronger capability for handling low contrast, scale variation, and noisy pavement backgrounds. These architectural components collectively contribute to more accurate crack localization and demonstrate higher methodological novelty and implementation-oriented detail, making the RL idea comparatively more impactful.

\end{tcolorbox}

\vspace{4pt}
\caption{Qualitative Comparison of Baseline, SFT, and RL Generated Research Ideas}
\label{tab:qualitative_comparison_3}

\end{table*}

\begin{table*}[t]
\centering
\small
\setlength{\tabcolsep}{4pt}
\renewcommand{\arraystretch}{1.12}

\begin{tcolorbox}[
enhanced,
width=\textwidth,
colback=white,
colframe=black,
fonttitle=\bfseries,
sharp corners,
boxrule=0.8pt
]

\textbf{Research Goal:}

\vspace{2pt}
The core objective of this research is to design a management system for equalizing charge among mixed power sources constituted by accumulators and supercapacitors, addressing the challenge of performance inconsistencies due to capacity differences among individual power sources. The charging system aims to facilitate rapid and consistent equalizing charging in series-parallel configurations while improving safety, longevity, and efficiency of the power pack.

\vspace{0.35cm}
\textbf{Color Legend:}

\vspace{2pt}
\baselinehl{Baseline} weak/abstract technical grounding \hspace{0.4cm}
\sfthl{SFT} moderate structured grounding \hspace{0.4cm}
\rlhl{RL} implementation-oriented grounding

\vspace{0.35cm}

\begin{tabularx}{\textwidth}{
>{\raggedright\arraybackslash}p{1.9cm}
>{\raggedright\arraybackslash}X
}
\toprule
\textbf{Method} & \textbf{Generated Idea Description} \\
\midrule

\textbf{Baseline} &
The proposed management system employs a feedback-based control mechanism to regulate the charging process of a mixed power source configuration comprising accumulators and supercapacitors. It utilizes voltage and current sensors to monitor the state of each individual power source in a series-parallel arrangement. These measurements are processed through a signal conditioning module to normalize and filter the input signals. A central control unit integrates the sensor data with predefined reference values to compute the required adjustments for balancing the charge distribution. The system dynamically adjusts the charging current applied to each power source based on its specific characteristics. \baselinehl{The control logic incorporates a loss function that quantifies the deviation between the actual and desired charge levels.} \baselinehl{This loss function is continuously updated using a feedback loop that compares the system's output against the target parameters.} The system operates in a closed-loop configuration where the output of the control unit directly influences the charging circuitry. \\
\midrule

\textbf{SFT} &
The proposed equalizing charging management system operates by utilizing two main charging modes, series and parallel, to effectively balance the charges across accumulators and supercapacitors in a series-parallel configuration. \sfthl{Initially, the system identifies the lowest voltage among all connected power sources, enabling the selection of a specific accumulator for equalizing charging through the series mode.} This involves sequentially connecting the selected accumulator in series with other elements while monitoring their voltages. \sfthl{As charging progresses, the system transitions to the parallel mode, wherein the supercapacitor is charged alongside the accumulator, ensuring that both components receive equivalent charge levels based on their respective capacities.} The design incorporates a control mechanism that dynamically adjusts the charging process according to real-time voltage measurements. \sfthl{The interaction between these components is facilitated through a feedback loop that continuously assesses the charging status and modifies the operational parameters accordingly.} \\
\midrule

\textbf{RL} &
The proposed equalizing charging management system integrates two main strategies: a parallel equalizing charging method and a series equalizing charging method. \rlhl{Initially, the parallel equalizing charging method employs a switching mechanism to charge all power sources simultaneously, utilizing a control unit that monitors voltage differences to determine when to initiate equalizing charge.} \rlhl{When voltage disparities exceed a predefined threshold, the system activates switches to connect capacitors in parallel, allowing for rapid equalization at high current levels without significant energy loss.} Subsequently, the series equalizing charging method sequentially charges each power source while monitoring their voltages to prevent overcharging. \rlhl{The architecture includes components such as a programmable logic controller (PLC), voltage detection circuits, and switching mechanisms to manage the flow of charge.} \rlhl{The system operation ensures that the charging current is maintained below a safe threshold, preserving the integrity of the supercapacitors and accumulators during equalizing processes.} \\
\bottomrule
\end{tabularx}

\par\vspace{0.35cm}
\noindent\textbf{Observation:}
The RL-generated idea demonstrates stronger translational impact readiness than the baseline and SFT outputs because it specifies executable control logic, concrete hardware-facing components, and explicit operational constraints. Relative to baseline abstraction and SFT-level structure, the RL output provides a clearer sensing-to-actuation pathway (voltage detection, threshold-triggered switching, and staged charging control), which improves implementation traceability and engineering testability. The inclusion of safety-relevant constraints (for example, controlled charging current and overcharge prevention logic) further increases deployment plausibility for real battery-management settings. Overall, the RL result is better positioned for practical prototyping due to higher mechanism concreteness, clearer decision flow, and tighter alignment with system-level operating requirements.

\end{tcolorbox}
\vspace{0.5cm}
\caption{Qualitative Comparison of Baseline, SFT, and RL Generated Research Ideas}
\label{tab:qualitative_comparison_4}

\end{table*}





\end{document}